\documentclass[preprint,5p,times,twocolumn]{elsarticle}

\usepackage{graphicx}%
\usepackage{epstopdf}
\usepackage{multirow}%
\usepackage{amsmath,amssymb,amsfonts}%
\usepackage{amsthm}%
\usepackage{mathrsfs}%
\usepackage[title]{appendix}%
\usepackage{xcolor}%
\usepackage{textcomp}%
\usepackage{manyfoot}%
\usepackage{booktabs}%
\usepackage{algorithm}%
\usepackage{algorithmicx}%
\usepackage{algpseudocode}%
\usepackage{listings}%
\usepackage{float}
\usepackage{subfigure}
\usepackage{bbm}
\usepackage[breaklinks=true]{hyperref}
\hypersetup{colorlinks = true, allcolors = blue}
\usepackage[nameinlink]{cleveref} 
\Crefformat{figure}{#2Fig.~#1#3}
\Crefmultiformat{figure}{Figs.~#2#1#3}{ and~#2#1#3}{, #2#1#3}{ and~#2#1#3}
\usepackage{caption}

\begin{document}

\captionsetup[figure]{labelfont={bf},labelformat={default},labelsep=period,name={Fig.}}
\captionsetup[table]{labelfont={bf}, labelsep=newline,font=footnotesize, justification=raggedright, singlelinecheck=false}

\begin{frontmatter}



\title{ECRTime: Ensemble Integration of Classification and Retrieval for Time Series Classification}


\author{Fan Zhao}

\author{You Chen\texorpdfstring{\corref{cor1}}{}}
\ead{ahjzcy@ahjzu.edu.cn}

\address{College of Environment and Energy Engineering, Anhui Jianzhu University, Hefei 230601, P. R. China}

\cortext[cor1]{Corresponding author}


\begin{abstract}
Deep learning-based methods for Time Series Classification (TSC) typically utilize deep networks to extract features, which are then processed through a
combination of a Fully Connected (FC) layer and a SoftMax function. However, we have observed the phenomenon of inter-class similarity and intra-class inconsistency in the datasets from the UCR archive and further analyzed how this phenomenon adversely affects the ``FC+SoftMax'' paradigm. To address the issue, we introduce ECR, which, for the first time to our knowledge, applies deep learning-based retrieval algorithm to the TSC problem and integrates classification and retrieval models. Experimental results on 112 UCR datasets demonstrate that ECR is state-of-the-art(sota) compared to existing deep learning-based methods. Furthermore, we have developed a more precise classifier, ECRTime, which is an ensemble of ECR. ECRTime surpasses the currently most accurate deep learning classifier, InceptionTime, in terms of accuracy, achieving this with reduced training time and comparable scalability.

\end{abstract}




\end{frontmatter}


\section{Introduction}
\label{sec1}

Time series data is extensively applied in various domains, including weather modeling, retail operations, financial forecasting, and many other sectors. This paper specifically focuses on the classification of univariate time series data. In academia, this field primarily consists of two methodological categories: distance-based and feature-based approaches. Distance-based methods, which serve as the foundational baseline in the field, classify data by computing similarities in the original raw time series, using pre-established distance metrics like Dynamic Time Warping (DTW) or Euclidean distance. DTW is notably effective in handling translational variances compared to the Euclidean distance. On the other hand, feature-based approaches extract feature vectors from the raw data and then employ classifiers such as Support Vector Machines (SVM), logistic regression, and decision trees to determine classification results.

Recent years have seen a growing body of research utilizing deep learning, especially deep convolutional network techniques, to address TSC challenges. Typically, these methods begin by extracting features using a deep network, and then proceed to classification through a FC layer combined with a SoftMax function. In the training phase, the FC layer fundamentally learns a weight matrix \({W_{d*c}}\)\, where \(d\) denotes the dimension of the feature vector, and \(c\) indicates the number of classes. This process is tantamount to learning a proxy(\(d*1\) dimension) for each class, leading to the convergence of features from a particular class near their respective proxy. During the testing phase, the classification is determined by applying the SoftMax to the distances between the test sequence features and the proxies for all classes.

\begin{figure}[htp]%
\centering
\includegraphics[width=0.48\textwidth]{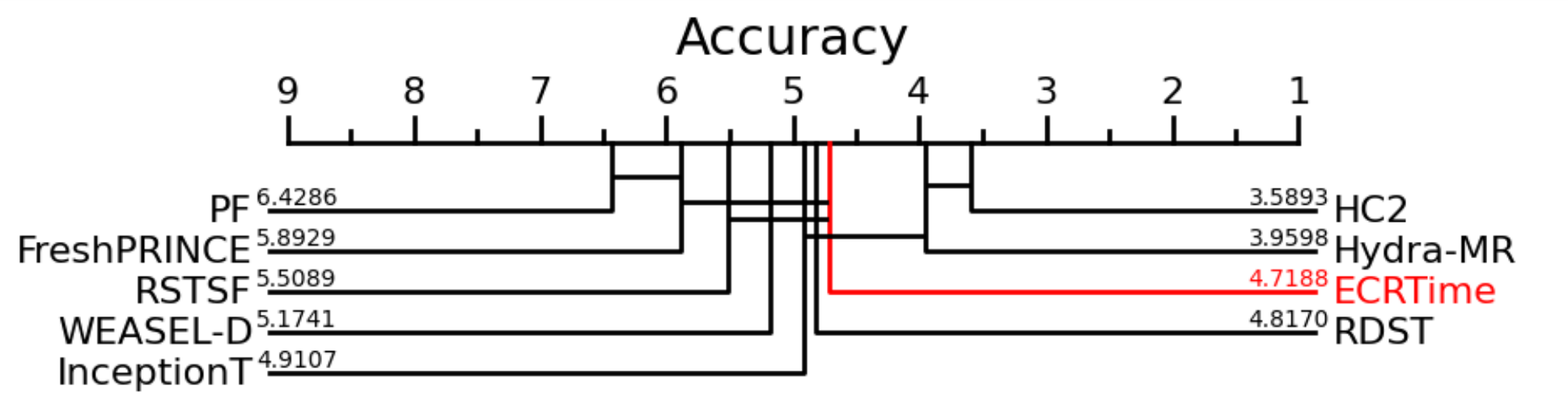}
\caption{Critical difference diagram showing the performance of ECRTime compared to the current SOTA classifiers on 112 datasets from the UCR archive.}
\label{ECRTime_vs_sotas}
\end{figure}

\begin{figure}[htbp]%
\centering
\includegraphics[width=0.48\textwidth]{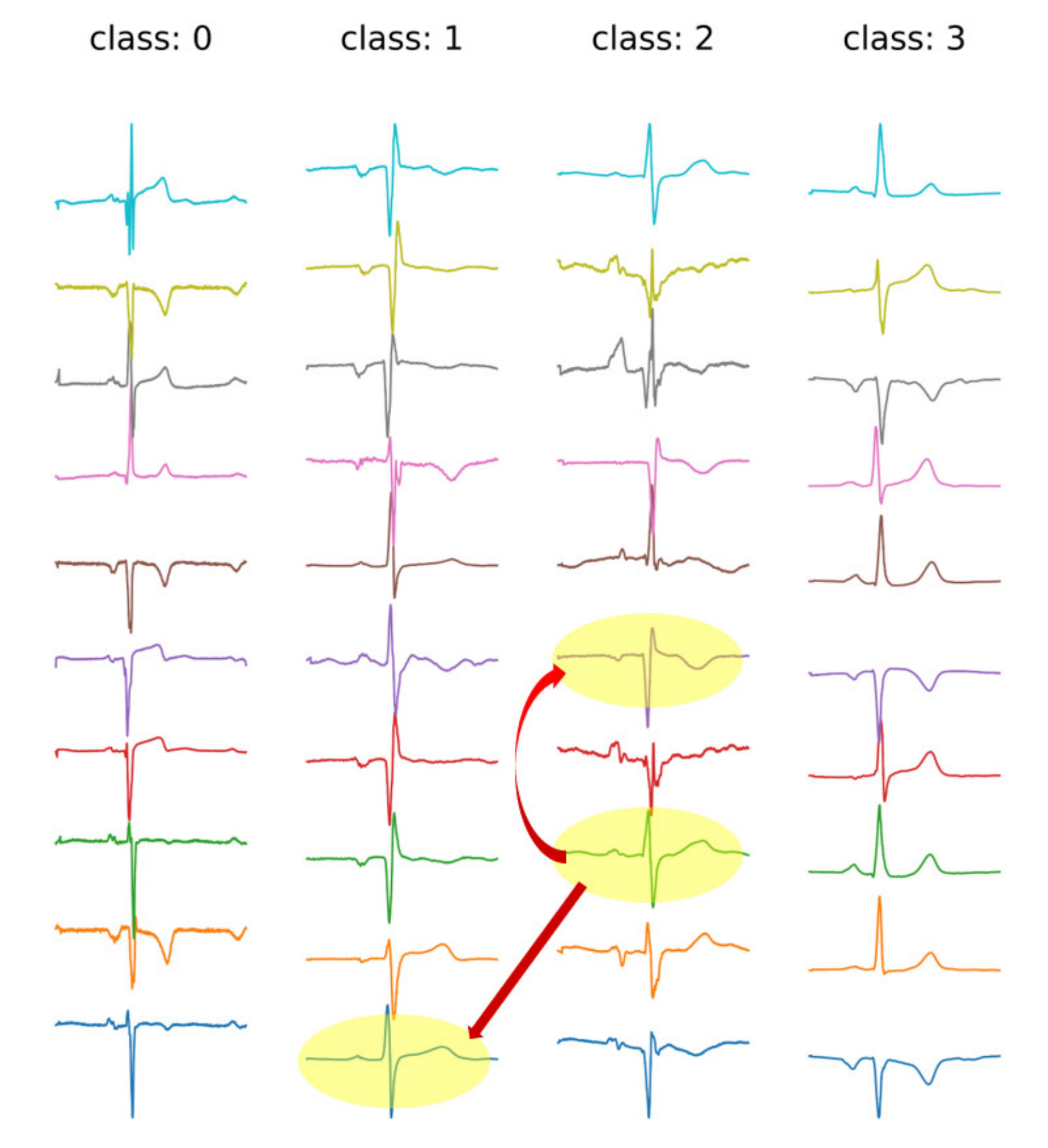}
\caption{Visualization of the CinCECGTorso dataset from the UCR archive, which comprises four categories. The display showcases the first 10 sequences sampled from each category.}
\label{CinCECGTorso_numclass4}
\end{figure}

\begin{figure}[htbp]%
\centering
\includegraphics[width=0.48\textwidth]{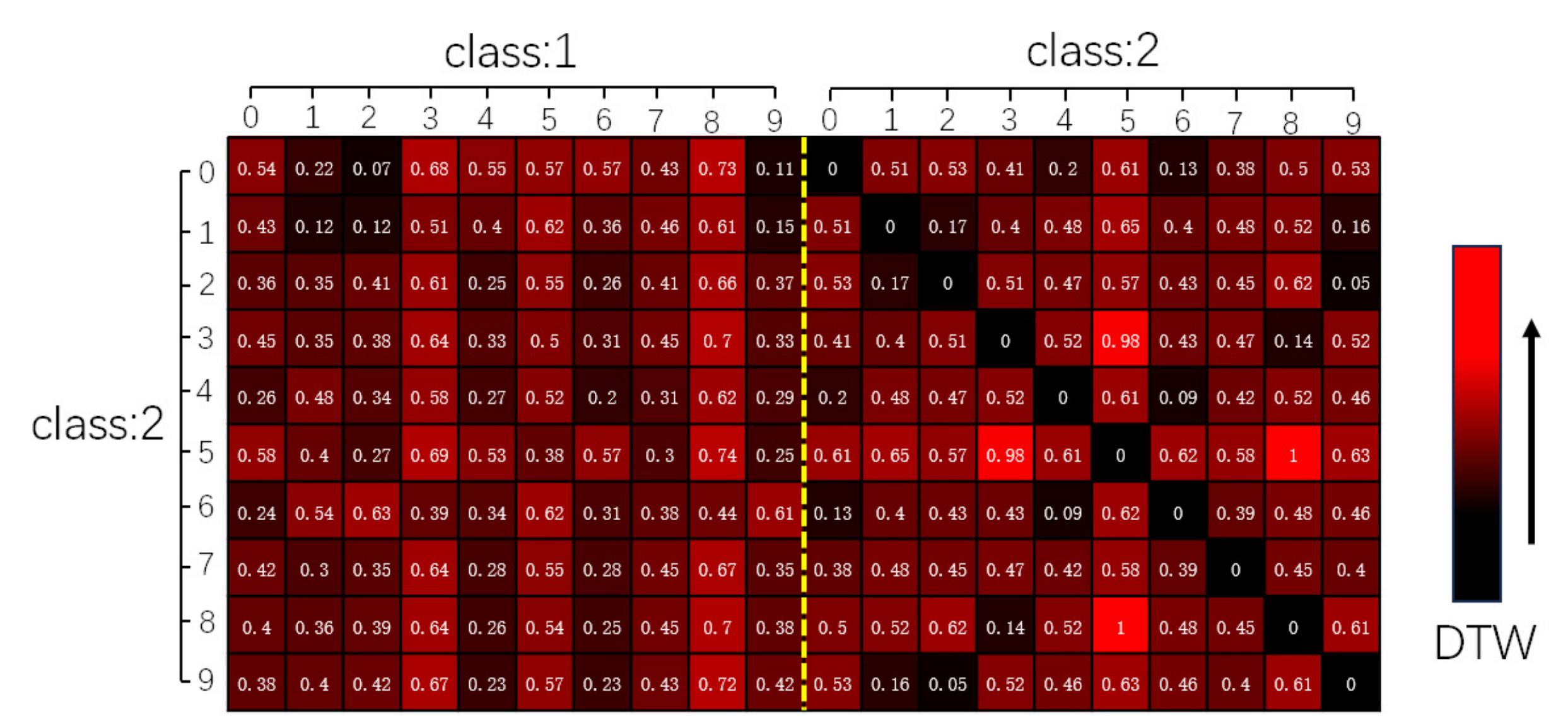}
\caption{Conducting a heatmap analysis based on DTW distance for series of class1 and class2 in the \Cref{CinCECGTorso_numclass4}.}\label{CinCECGTorso_HotMap}
\end{figure}

We also conducted an analysis of datasets commonly used in TSC from the UCR archive\cite{bib42}, and we found that the data exhibits the phenomenon of inter-class similarity and intra-class inconsistency. \Cref{CinCECGTorso_numclass4} presents a visual analysis employing the CinCECGTorso dataset as an illustrative example. Examining two sequence pairs highlighted with red arrows, it is apparent that sequences from class1 and class2 share similarities in shape, yet there is a clear distinction within class2 itself. Owing to the robustness of Dynamic Time Warping (DTW) distance against shifts, it more precisely captures sequence discrepancies. Consequently, for the sequences sampled in \Cref{CinCECGTorso_numclass4}, a heatmap based on DTW distance was created, as depicted in \Cref{CinCECGTorso_HotMap}. This heatmap reveals that the left side has numerous smaller distance values compared to the right side, suggesting similarities between sequences in class2 and class1, while also indicating varying patterns within class2.

\begin{figure}[htbp]%
\centering
\includegraphics[width=0.48\textwidth]{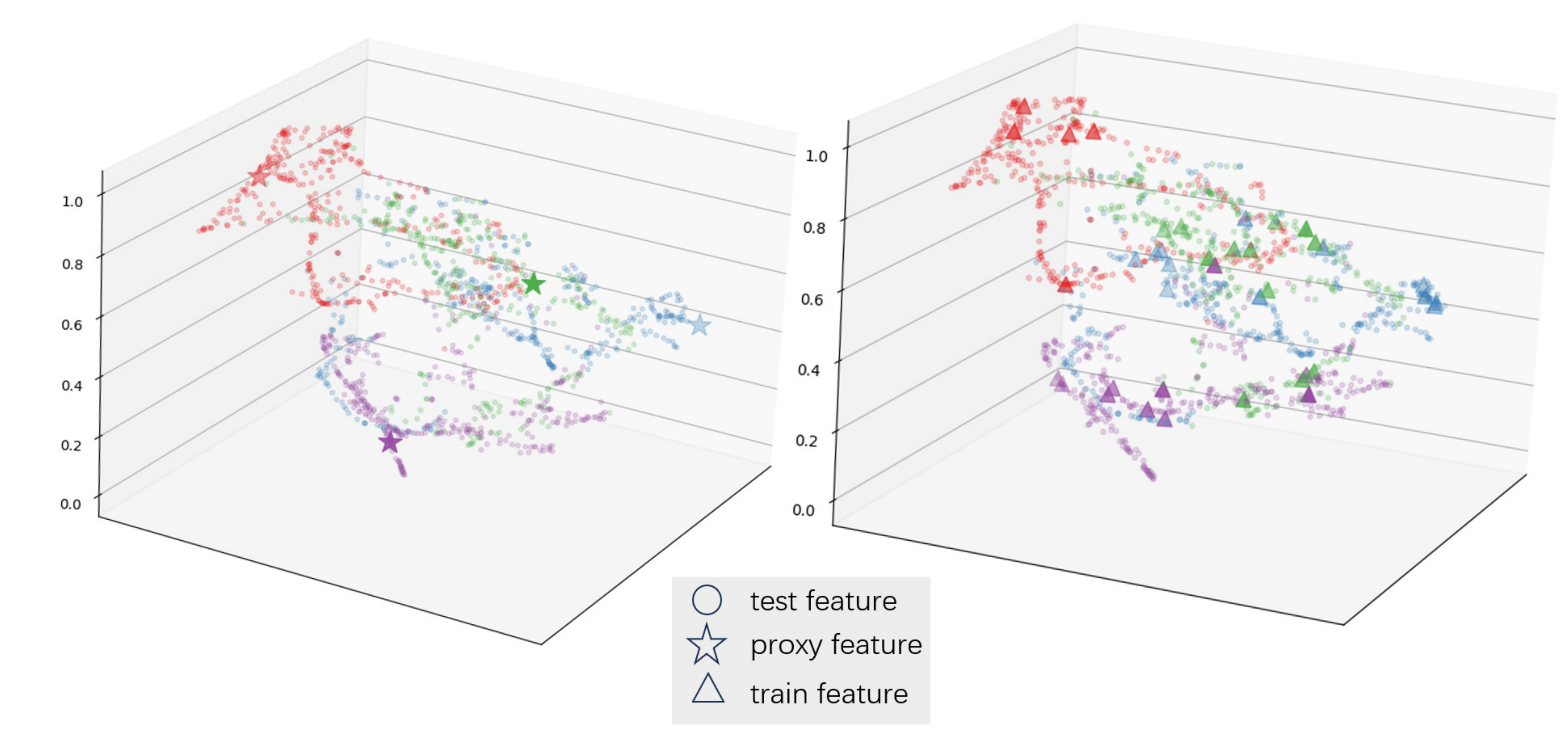}
\caption{Within the framework of the ``FC+SoftMax'' classification paradigm, the distribution of feature vectors for the CinCECGTorso dataset is visualized using t-SNE. The four colors in the figure represent four classes. The left graph represents the distribution of features from the test set and proxy features, while the right graph illustrates the distribution of features from the training set (used as a reference library) and features from the test set.}\label{tsne}
\end{figure}

Further, the aforementioned phenomenon will impede the resolution of TSC using the ``FC+SoftMax'' paradigm. On one hand, it requires broader representation of each class through proxies; on the other hand, representing a class with a single feature in this paradigm often leads to overfitting. Given the similar distribution of the training and test sets, a viable solution to this issue is utilizing more detailed and inclusive features from the training set as new ``proxies''. To validate this, in \Cref{tsne}, we present the distribution of feature vectors across all classes, alongside their respective proxies and training set features. It is evident that the coverage provided by the training set features is more extensive than that by old proxies.

In our approach, we initially drew inspiration from \cite{bib1} to design a ResNet-type network as the backbone for feature extraction. Subsequently, we constructed a comparison library utilizing the features from all sequences in the training set and substituted the SoftMax classifier with a 1-NN classifier. This method represents each category by the collective features of its corresponding sequences in the training set, rather than relying on a single proxy feature. The label for a test sample is determined by the most similar sequence in the library. Experimental evidence indicates that this strategy has enhanced classification performance. Moreover, the ``FC+SoftMax'' paradigm typically utilizes the CrossEntropy loss function during training. This function's objective is to align the features of each class's sequences more closely with their respective proxies. The 1-NN classifier functions by identifying the most similar target in the library, assigning its label as the predicted label. Consequently, the training goal of the CrossEntropy loss function does not explicitly align with the prediction objective of the 1-NN classifier. To better tailor the loss function to the classifier's needs, we incorporated a deep learning-based retrieval method, conceptualizing class as 'id' in retrieval tasks. In the training process, we employed the hard triplet loss, which effectively narrows the feature distances within the same class while widening those between different classes. Unfortunately, triplet loss cannot measure the overall spatial distribution of features, while the CrossEntropy loss does not have enough discriminant between features. Consequently, our experiments demonstrated that the retrieval model outperformed the classification model in certain datasets, yet was less effective in others. 

To harness the advantages of both methods, we integrate classification and retrieval models to develop the ECR model. Experimental validation on 112 UCR UTSC problems has demonstrated that ECR attains a SOTA status among deep learning-based approaches. Furthermore, by integrating three ECR models in a manner similar to InceptionT\cite{bib13}, we crafted the more refined ECRTime model, as depicted in \Cref{ECRTime_pipeline}. This advanced model not only outperforms InceptionT but also surpasses most contemporary SOTA methods in performance. For a detailed comparison, see \Cref{ECRTime_vs_sotas}. In summary, the main contributions of this paper can be summarized as follows:

\begin{enumerate}
\item[--] Explored the phenomenon of inter-class similarity and intra-class inconsistency across datasets from the UCR archive, and analyzed how the commonly used ``FC+SoftMax'' combination in classification tasks is adversely affected by it.

\item[--] To address the aforementioned issue, the SoftMax classifier is replaced with a 1-NN classifier, and for the first time, a retrieval method based on deep learning is introduced to explicitly align the training loss function with the 1-NN classification objective.

\item[--] To achieve optimal classification performance, we initially combine classification and retrieval models to develop the ECR model. This method was further improved by integrating three ECR models, resulting in the development of the ECRTime model. Comprehensive testing on 112 UCR datasets has shown that ECRTime outperforms other state-of-the-art (SOTA) methods in Time Series Classification (TSC).
\end{enumerate}

The structure of the remaining content of this paper is as follows: Section~\ref{sec2} introduces related work. In Section~\ref{sec3}, we first present the overall algorithm process, then provide detailed descriptions of key modules, including: Backbone network, Loss function, Distance and Ensemble module. In Section~\ref{sec4}, we first introduce the experimental setup of ECRTime, then compare it with other SOTA methods on 112 datasets of the UCR archive, followed by sensitivity study. Finally, Section~\ref{sec5} provides the conclusion of this paper along with the plan for future work.

\section{Related work}\label{sec2}

\subsection{State-Of-The-Art Time Series Classifiers}\label{subsec2.1}

The time series classification (TSC) problem represents a foundational challenge within the domain, with the academic community having proposed a multitude of effective algorithms to contend with this intricacy. Backoff-2023\cite{bib27} summarizes the current eight state-of-the-art (SOTA) Time Series Classification (TSC) classifiers, which are HIVE-COTE 2.0\cite{bib11}, Hydra-MR\cite{bib28}, InceptionT\cite{bib13}, RDST\cite{bib29}, WEASEL-D\cite{bib30}, RSTSF\cite{bib31}, FreshPRINCE\cite{bib32} and PF\cite{bib33}. We categorize them into ensemble-based and feature-based methods. Among these, the most accurate ensemble method is HIVE-COTE 2.0, which is also currently the best time series classifier. It employs more advanced ensemble techniques compared to its predecessor, Hive-COTE 1.0\cite{bib16}. It includes the following: STC, Temporal Dictionary Ensemble (TDE)\cite{bib20}, Diverse Representation Canonical Interval Forest (DrCIF), and Arsenal. Among these methods, DrCIF, developed by the authors of this paper, extends the Canonical Interval Forest (CIF) \cite{bib21}. Meanwhile, Arsenal, an ensemble of compact ROCKET classifiers, generates valuable probability values for each class during predictions using CAWPE. Presently, these ensemble methods represent the state-of-the-art in time series classification. Nevertheless, they typically exhibit high time complexity, posing substantial challenges for practical application. In our research, we employed an ensemble approach that balances high accuracy with comparatively lower time complexity.

Hydra-MR stands as the most accurate feature-based method to date, achieving an excellent balance between accuracy and time consumption. It amalgamates the Hydra algorithm\cite{bib28} with the MR (MultiRocket) model\cite{bib2}. Building on the foundational work of Rocket\cite{bib5} and MiniRocket\cite{bib6}, MultiRocket introduces a variety of pooling operations and transformations, enhancing the diversity of feature distributions. This advancement not only boosts classification accuracy but also maintains computational efficiency. HYDRA, a dictionary-based method, transforms input time series data using a collection of randomly selected convolutional kernels grouped together. It quantifies the frequency of kernels that most closely match the input time series at each point in time. These quantifications are then utilized to train a linear classifier. 

\subsection{Deep learning-based methods}\label{subsec2.2}

The current academic emphasis on resolving TSC challenges predominantly resides in the domain beyond deep learning, where non-deep learning methods prevail in terms of both prevalence and performance. DL-review\cite{bib34}, as an influential work in the field of Time Series Classification (TSC) within the deep learning domain, summarizes nine advanced deep learning-based time series classifiers, including: Resnet\cite{bib1}, FCN\cite{bib1}, Encoder\cite{bib35}, MLP\cite{bib1}, Time-CNN\cite{bib36}, TWISEN\cite{bib37}, MCDCNN\cite{bib38}, MCNN\cite{bib12}, and t-LeNeT\cite{bib39}. Among these, Resnet and FCN exhibit relatively optimal performance. Resnet consists of three residual blocks, each followed by a Global Average Pooling (GAP) layer and a softmax classifier at the end. The number of neurons in the classifier corresponds to the number of classes in the dataset. Within each residual block, three convolutions are initially performed, the output of which is added to the block's input before being passed to the subsequent layer. FCN comprises three convolutional blocks, each containing three sequential operations: a convolution, batch normalization, and a ReLU activation function. The output of the third convolutional block undergoes averaging across the entire time dimension, forming the Global Average Pooling (GAP) layer. Subsequently, a conventional softmax classifier is connected to the output of the GAP layer. Subsequently, it was found in the \cite{bib40} that directly ensembling these deep learning models could further enhance the algorithm's performance. Based on this discovery, InceptionT\cite{bib13} was initially inspired by the Inception-v4\cite{bib41} network from computer vision tasks and designed the ``AlexNet'' of TSC - the Inception net. It then ensembled these five models, ultimately achieving an accuracy on UCR85 comparable to HIVE-COTE 1.0. Moreover, Backoff-2023\cite{bib27} comprehensively reviews and summarizes numerous deep learning models in current TSC tasks, ultimately concluding that InceptionT is currently the best deep learning-based time series classifier.

\section{Method}\label{sec3}

In this section, we present the overall network framework of the ECR module, which serves as the ensemble component of ECRTime. Subsequently, detailed descriptions of each module within the network are provided. Ultimately, the final result is generated through a straightforward but effective ensemble strategy.

\subsection{ECR Framework}\label{subsec3.1}

\begin{figure*} [ht]
    \centering
    \subfigure[Training pipeline]{
        \label{train}
        \includegraphics[width=0.8\linewidth]{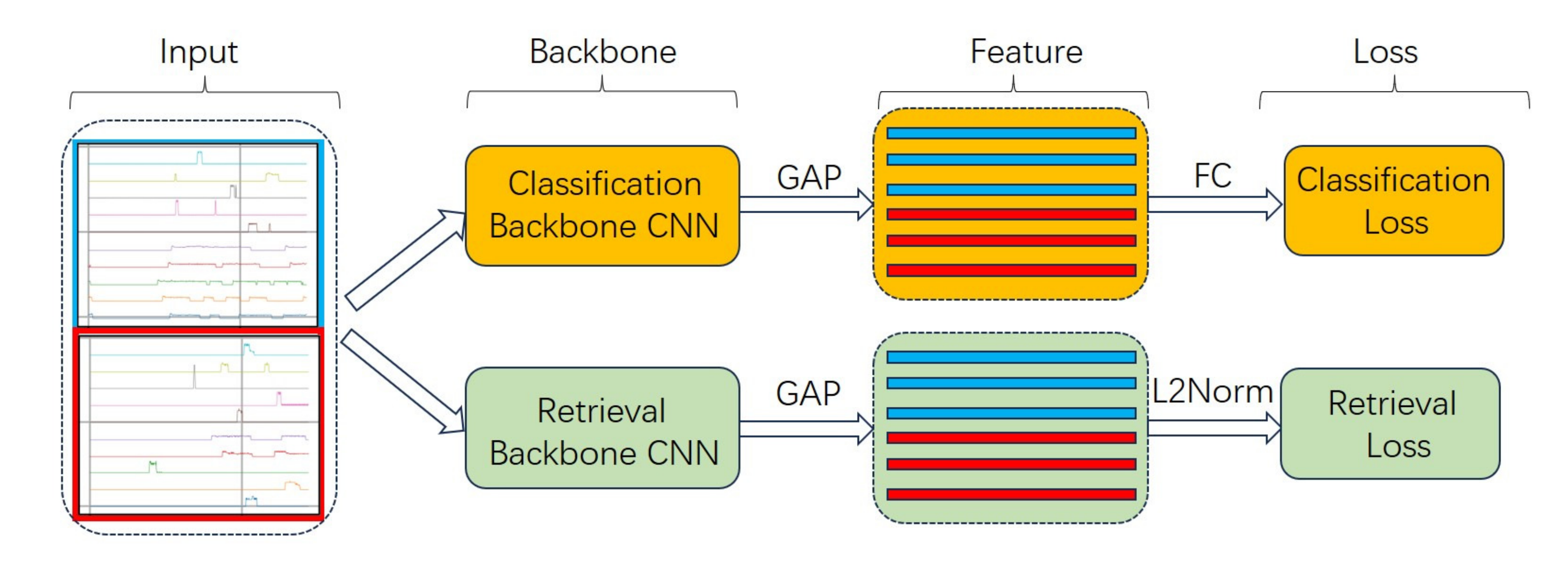}}
    \subfigure[Inference pipeline]{
        \label{inference}
        \includegraphics[width=0.8\linewidth]{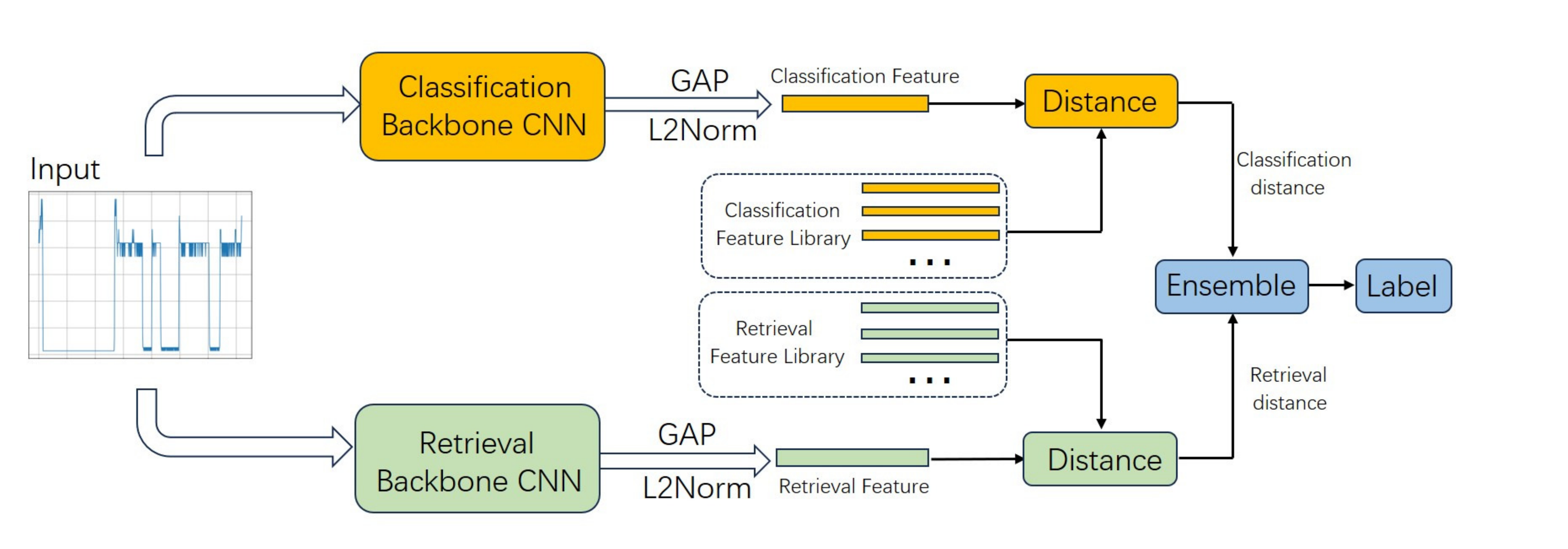}}
    \caption{Training pipeline(top) and inference pipeline(bottom) of the ECR. GAP: Global Average Pooling, FC: Fully Connected Layer, L2Norm: L2 normalization. In the top subplot, the blue and red boxes represent time series of different categories.}
    \label{ECR_pipeline}
\end{figure*}

\Cref{ECR_pipeline} illustrates the overall structure of the ECR, which includes two stages: training and inference, as shown in \Cref{train} and \Cref{inference} respectively.

We use \({x} \in {R^{1 \times L}}\) to represent a single time series, \(1\) to indicate that it is univariate, and \(L\) to represent the length of the series. During the training phase, each mini-batch input is denoted as \(X\), the input label set as \(Y\), and the batch size as \(B\). Each mini-batch includes \(C\) different categories, and each category contains \(m\)  samples, thus resulting in \(B = C \times m\), \(X \in {R^{B \times 1 \times {\rm{L}}}}\), \(Y \in {\{ 1,2,...,C\} ^B}\).

As shown in \Cref{train}, the training framework consists of two independent forward branches: one trained based on classification method and the other based on retrieval method, both sharing the same input \(X\). During training, \(X\) passes through the classification backbone and retrieval backbone to extract feature vectors \({F_{Cls}} \in {R^{B \times d \times L}}\) and \({F_{Ret}} \in {R^{B \times d \times L}}\), respectively. Note that the two backbones have the same structure but do not share weights. \(d\) represents the number of channels in the feature vector. Subsequently, \({F_{Cls}}\) and \({F_{Ret}}\) are both dimensionally reduced to \({{F}_{Cls}^{'}} \in {R ^ {B \times d}}\) and \(F_{Ret}^{'} \in {R^{B \times d}}\) through Global Average Pooling(GAP) on the \(L\) dimension. In the classification branch, \({F}_{Cls}^{'}\) is followed by a Fully Connected Layer, and the output is then fed into the classification loss for learning. In the retrieval branch, \(F_{Ret}^{'}\) , after undergoing L2 norm operation, is input into the retrieval loss.

In the inference phase shown in \Cref{inference}, for the test sequence, features are extracted based on the backbone networks in the two branches, and then both are reduced in dimension and normalized through Global Average Pooling and L2 Normalization. Additionally, we will pre-extract features of all sequences in the training set in this manner and construct both a classification feature library and a retrieval feature library. Subsequently, the classification and retrieval features of the test sequence are compared with corresponding library features in the Distance module. If there are \(N\) features in the library, \(1 \times N\) classification distance vectors and retrieval distance vectors are outputted respectively. Finally, the predicted category is outputted by ensembling these two types of distances. In the following sections, we will specifically introduce the key modules in the training and inference, including: Backbone, Loss, Distance and Ensemble.

\subsection{Backbone network structure}\label{subsec3.2}

\begin{figure}[htp]%
\centering
\includegraphics[width=0.48\textwidth]{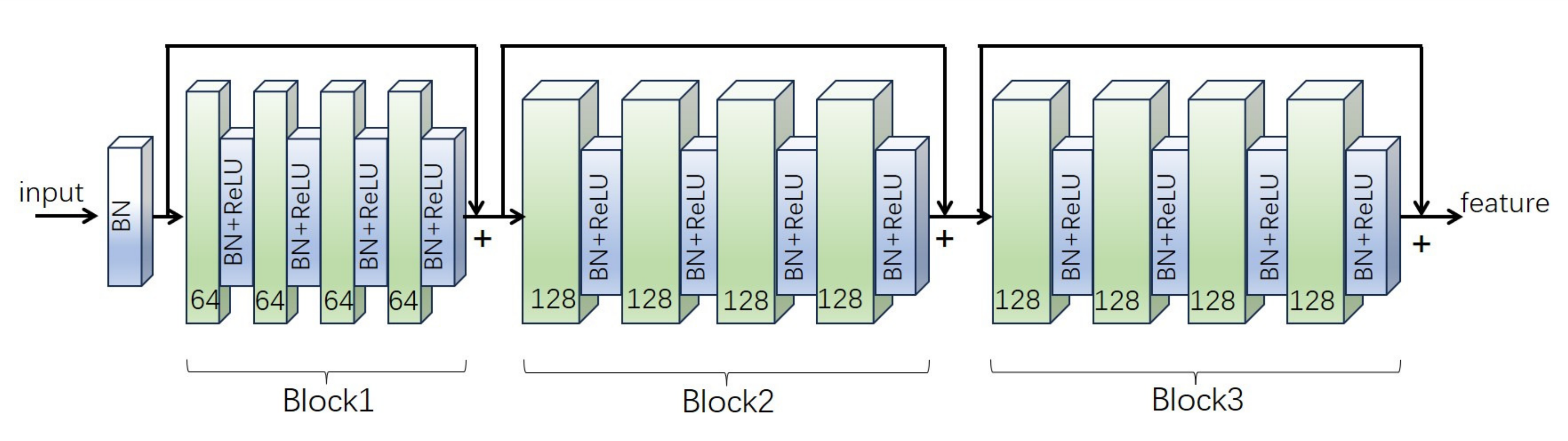}
\caption{Backbone network structure.}\label{Backbone}
\end{figure}

During these years of rapid development in deep learning, many backbone networks for feature extraction have emerged, yet ResNet\cite{bib22}, based on residual connections, remains the preferred choice in numerous application scenarios. The residual structure, without introducing additional parameters and computational load, can effectively mitigate gradient vanishing on one hand, ensuring the continuity of parameter learning; on the other hand, deeper networks can be constructed based on residual connections, and generally speaking, the deeper the network, the stronger its feature extraction capability. Existing works such as \cite{bib1, bib23, bib24} have already applied ResNet in the TSC field. Based on the structures validated in these methods, we designed the backbone module of this paper, as shown in \Cref{Backbone}.

\begin{table}[htp]
\caption{The parameters of each layer in block1 of the backbone network.}
    \centering
    \begin{tabular}{|c|c|c|}
    \hline
        \textbf{type} & \textbf{kernel size/stride/pad } & \textbf{input size} \\ \hline
        conv & 9/1/4 & \(1 \times 1460\) \\ \hline
        BN+ReLU & - & \(64 \times 1460\) \\ \hline
        conv & 7/1/3 & \(64 \times 1460\) \\ \hline
        BN+ReLU & - & \(64 \times 1460\) \\ \hline
        conv & 5/1/2 & \(64 \times 1460\) \\ \hline
        BN+ReLU & - & \(64 \times 1460\) \\ \hline
        conv & 3/1/1 & \(64 \times 1460\) \\ \hline
        BN+ReLU & - & \(64 \times 1460\) \\ \hline
    \end{tabular}
    \label{Backbone_tab}
\end{table}

Similar to the structure in \cite{bib1}, the input in \Cref{Backbone} first goes through a BN (Batch Normalization) layer for normalization, then through three residual blocks for feature extraction. We also follow the approach in \cite{bib23} to convert each layer to 1D form, including the convolutional layers and BN layers. The 1D structure is more suitable for one-dimensional time series input, requires less padding, and reduces computational complexity. Furthermore, we adopt a (9, 7, 5, 3) kernel size combination, in contrast to the (8, 5, 3) configuration cited in \cite{bib1,bib23}. This choice is due to the general preference for odd-sized kernels in convolutional networks, which minimize layer distortion and ensure symmetry around the output pixel in the preceding layer. In \Cref{Backbone_tab}, we use a sequence with an input size of \(1 \times 1460\) as an example to show the specific parameters of each layer in the first block. The other two blocks have similar parameter formats, differing only in the channel dimension.

\subsection{Loss}\label{subsec3.3}

As shown in \Cref{train}, the training phase employs two types of loss functions: the classification branch utilizes cross-entropy loss, denoted as \({L_{Cls}}\), and the retrieval branch employs the hard version of triplet loss, denoted as \({L_{Ret}}\). Based on the symbol definitions in Section~\ref{subsec3.1}, \({L_{Cls}}\) is defined in the following form:

\begin{equation}
{L_{Cls}} = ( - \frac{1}{B})\sum\limits_{i = 1}^B {[\sum\limits_{j = 1}^C \mathbbm{1}( {y^i} = j) log\frac{{{{\rm{e}}^{z_j^i}}}}{{\sum\limits_{k = 1}^C {{e^{z_k^i}}} }}]} 
\label{eq1}
\end{equation}

In Eq.~\eqref{eq1}, \(z\) represents the vector obtained after \({{F}_{Cls}^{'}}\) passes through the FC layer, and the indicator function \(\mathbbm{1}( {y^i} = j) \) is defined as follow: 

\begin{equation}
\mathbbm{1}( {y^i} = j)  = \left\{ {\begin{array}{*{20}{c}}
{1,{y^i} = j}\\
{0,{y^i} \ne j}
\end{array}} \right.
\label{eq2}
\end{equation}

Based on Eq.~\eqref{eq2}, Eq.~\eqref{eq1} can be further simplified to the following form:

\begin{equation}
{L_{Cls}} = ( - \frac{1}{B})\sum\limits_{i = 1}^B {log\frac{{{{\rm{e}}^{z_{{y^i}}^i}}}}{{\sum\limits_{k = 1}^C {{e^{z_k^i}}} }}} 
\label{eq3}
\end{equation}

For retrieval tasks, triplet loss\cite{bib26} is generally used. It originally works on an anchor series \(A\), a positive sample \(P\) from the same class and a negative sample \(N\) from a different class. The objective is to minimize the distance between \(A - P\), while push away the \(N\). The formula of triplet loss is as follow:

\begin{equation}
{L_{Triplet}} = {[\frac{1}{{{H_P}}}\sum\limits_{i = 1}^{{H_P}} {{{\left\| {{g_A} - g_P^i} \right\|}_2}}  - \frac{1}{{{H_N}}}\sum\limits_{j = 1}^{{H_N}} {{{\left\| {{g_A} - g_N^j} \right\|}_2} + \alpha } ]_ + }
\label{eq4}
\end{equation}

Eq.~\eqref{eq4} iterates over and calculates each sample in the mini-batch input, taking the average as the final loss. In this context, \(g\) denotes the vector after L2 norm, \(H\) indicates the count of positive or negative samples. The term \(\alpha\) represents the margin between positive and negative samples. Additionally, the subscripts \(P\) and \(N\) correspond to positive and negative samples, respectively. The original triplet loss introduces many easily satisfied triplets, which lack contribution to the training, leading to slower and less efficient convergence. Therefore, this paper follows the approach in \cite{bib26} and uses hard triplet loss for training. This loss narrows the distance between the anchor sample and the farthest positive, while increasing the distance between the anchor and the closest negative, defined as follow:

\begin{equation}
{L_{Ret}} = {[\mathop {\max }\limits_{i \in [0,{H_P})} ({\left\| {{g_A} - g_P^i} \right\|_2}) - \mathop {\min }\limits_{j \in [0,{H_N})} ({\left\| {{g_A} - g_N^j} \right\|_2}) + \alpha ]_ + }
\label{eq5}
\end{equation}

\subsection{Distance}\label{subsec3.4}

In the training phase, two pivotal modules, the Backbone and the Loss, have been delineated earlier. Subsequently, in the testing phase, we elucidate the Distance module, conceived on the 1-NN classifier paradigm. As depicted in the testing procedure (\Cref{inference}), each time series input \({{x}^i}\) undergoes feature extraction via the classification and retrieval backbone, followed by processing through GAP and L2Norm, yielding output vectors \(f_{Cls}^i\) and \(f_{Ret}^i\). Concurrently, features are extracted from every sequence in the training set in a similar fashion, leading to the formation of two libraries: the classification library \({F_{Cls\_lib}} = \{ j \in [0,N)|f_{Cls}^j\} \) and the retrieval library \({F_{Ret\_lib}} = \{ j \in [0,N)|f_{Ret}^j\} \), where \(N\) signifies the total number of sequences in the training set.

In the Distance module, by iterating and calculating the Euclidean distance between \(f_{Cls}^i\) and all features in the classification library \({F_{Cls\_lib}}\), we obtain \(D(f_{Cls}^i,{F_{Cls\_lib}}) = {\{ D_{Cls}^1,D_{Cls}^2,...,D_{Cls}^N\} ^N}\), representing the distance between \(f_{Cls}^i\) and \({F_{Cls\_lib}}\). The distance between vector \(f_{Ret}^i\) and the retrieval library \({F_{Ret\_lib}}\) is denoted as \(D(f_{Ret}^i,{F_{Ret\_lib}}) = {\{ D_{Ret}^1,D_{Ret}^2,...,D_{Ret}^N\} ^N}\), calculated in a similar manner.

\subsection{Ensemble}\label{subsec3.5}

\begin{figure}[htp]%
\centering
\includegraphics[width=0.4\textwidth]{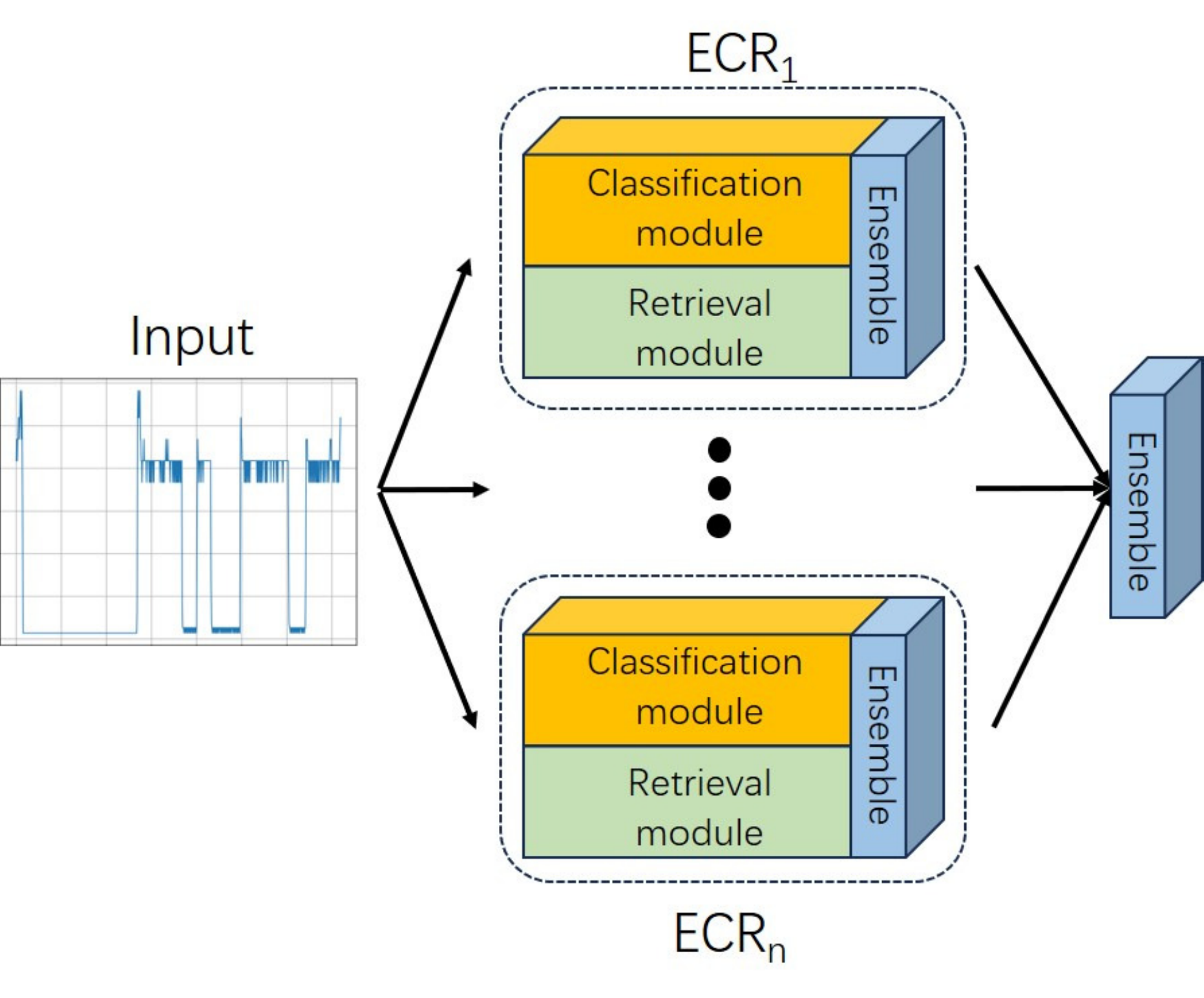}
\caption{Pipeline of ECRTime.}
\label{ECRTime_pipeline}
\end{figure}

Following the Distance module is the Ensemble module, and this paper involves two stages of ensemble operations, as illustrated in \Cref{ECRTime_pipeline}. The first stage ensembles classification and retrieval models to obtain ECR, and the second stage ensembles multiple ECRs to derive the final ECRTime model. In the first ensemble, as shown in \Cref{inference}, we average all corresponding elements in sets \(D(f_{Cls}^i,{F_{Cls\_lib}})\) and \(D(f_{Ret}^i,{F_{Ret\_lib}})\) to obtain the final distance set \(D({f^i},{F_{lib}}) = {\{ (D_{Cls}^1 + D_{Ret}^1)/2,(D_{Cls}^2 + D_{Ret}^2)/2,...,(D_{Cls}^N + D_{Ret}^N)/2\} ^N}\). Finally, the category of the corresponding sequence in the library, predicted for the test sequence \({x^i}\), is identified by taking \(\arg \min ( * )\) of \(D({f^i},{F_{lib}})\).

It is important to note that L2 normalization is applied to the classification features during testing to ensure uniformity in the value ranges of \(D(f_{Cls}^i,{F_{Cls\_lib}})\) and \(D(f_{Ret}^i,{F_{Ret\_lib}})\). This uniformity is crucial as it prevents the averaging of prediction results from being skewed by differing value ranges. Additionally, for two vectors with an L2 norm of 1, their Euclidean distance can be reformulated as \(\sqrt {2(1 - \cos \theta )} \) where \(\theta \) is the angle between the vectors. This ensures that both parties being averaged have the same value range, which is \([0,2]\).

\begin{equation}
{\widehat y_{i,c}} = \frac{1}{n}\sum\limits_{j = 1}^n {{\sigma _c}({x_i},{\theta _j})|{\forall _c} \in [1,C]}
\label{eq6}
\end{equation}

Furthermore, in the second ensemble, based on Eq.~\eqref{eq6}, multiple ECRs are integrated to obtain the final ECRTime model presented in this paper. In the formula, \(\widehat y_{i,c}\) represents the ensemble's output probability that the input time series, \(x_i\) belongs to class \(c\), This is equivalent to the average logistic output \(\sigma _c\)  across \(n\) randomly initialized ECRs.

\section{Experiments and results}\label{sec4}

\subsection{Experiment setup}\label{subsec4.1}

In this section, we evaluate the ECR and ECRTime on 112 datasets in the UCR univariate time series archive. ECRTime refers to an ensemble of three ECR modules, while the “ECRTime(n)” notation is used to denote an ensemble of n ECR modules. ``UCR112'' is used to denote the 112-version of the UCR archive in the following text. The experimental results are available on the website\footnote{\url{https://anonymous.4open.science/r/ECRTime-3834}}. Initially, the experimental setup is introduced.

\textbf{Datasets and SOTAs}: To compare with numerous advanced algorithms while avoiding excessively time-consuming experiments, we followed the approach used in \cite{bib11,bib27}, conducting experimentation with the 112 equal length problems in the 2019 version of the UCR archive. The comparison includes classifiers based on deep learning summarized in DL-review\cite{bib34} and state-of-the-art classifiers in the field of Time Series Classification(TSC) compiled in Backoff-2023\cite{bib27}. The corresponding comparison results are respectively sourced from \footnote{\url{https://github.com/hfawaz/dl-4-tsc/blob/master/results/results-ucr-128.csv}} and \footnote{\url{https://github.com/time-series-machine-learning/tsml-eval/tree/main/tsml_eval/publications/y2023/tsc_bakeoff/results}}.

\textbf{Configuring ECRTime}: Since the datasets in the UCR only consist of training and test sets, lacking a validation set, it is not possible to tune hyperparameters such as epochs. Therefore, we refer to the ResNet classifier in DL-review\cite{bib34} for hyperparameter settings. During the training phase, we set each mini-batch input to contain 4 categories, with 4 samples per category, resulting in a batch size of 16. The margin in the hard triplet loss is set to 0.1, and the optimizer used is Adam. For the classification branch, the learning rate is set at 1e-3, and for the retrieval branch, the learning rate is set at 1e-4. The scheduler used is ReduceLROnPlateau from PyTorch, the number of training epochs is set to 1500. Experimental environment configuration: PyTorch 2.0, Python 3.9.

\subsection{Comparing with deep learning-based methods}\label{subsec4.2}

\begin{figure}[htp]%
\centering
\includegraphics[width=0.48\textwidth]{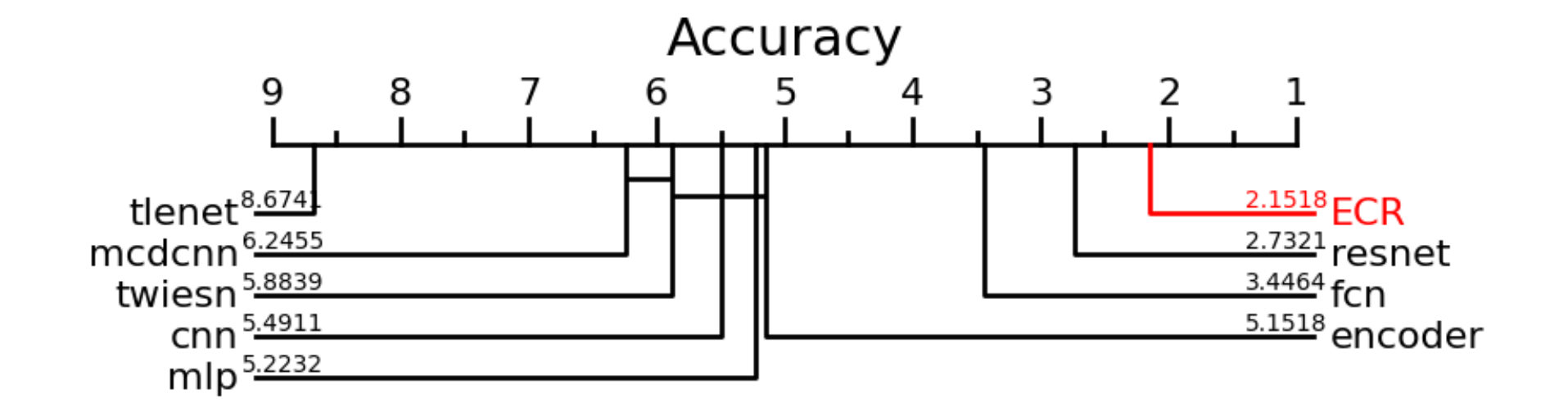}
\caption{Mean rank of ECR in terms of accuracy versus other deep learning methods on 112 datasets from the UCR archive.}\label{ECR_vs_DLs}
\end{figure}

To facilitate a comparison with a range of deep learning classifiers, we evaluated ECR against the eight methods detailed in DL-review\cite{bib34}. For result validation, we adhered to DL-review's methodology, training ECR for five iterations on UCR112. During these iterations, only the random seed varied, while the model's structure and training hyperparameters remained unchanged. The final reported accuracy represents the mean of these iterations. \Cref{ECR_vs_DLs} features a critical difference diagram that illustrates the accuracy comparisons between ECR and various deep learning models. The horizontal thick line across different models indicates no significant difference between them (p-value\textgreater 0.05). It is noted that ECR significantly outperforms all methods depicted in the figure (p-value\textless 0.05), including ResNet and FCN, which were identified as the most precise deep learning classifiers in DL-review at that time. In the subsequent Section~\ref{subsubsec4.5.5}, the discussion will focus on how ECRTime, an ensemble of ECR, markedly surpasses ECR, indicating that ECRTime also surpasses the aforementioned deep learning-based methods. Consequently, to reduce the redundancy of the experiments, we refrained from conducting a parallel comparative analysis for ECRTime as in \Cref{ECR_vs_DLs}.

\subsection{Comparing with SOTAs}\label{subsec4.3}

\begin{figure}[ht]%
\centering
\includegraphics[width=0.48\textwidth]{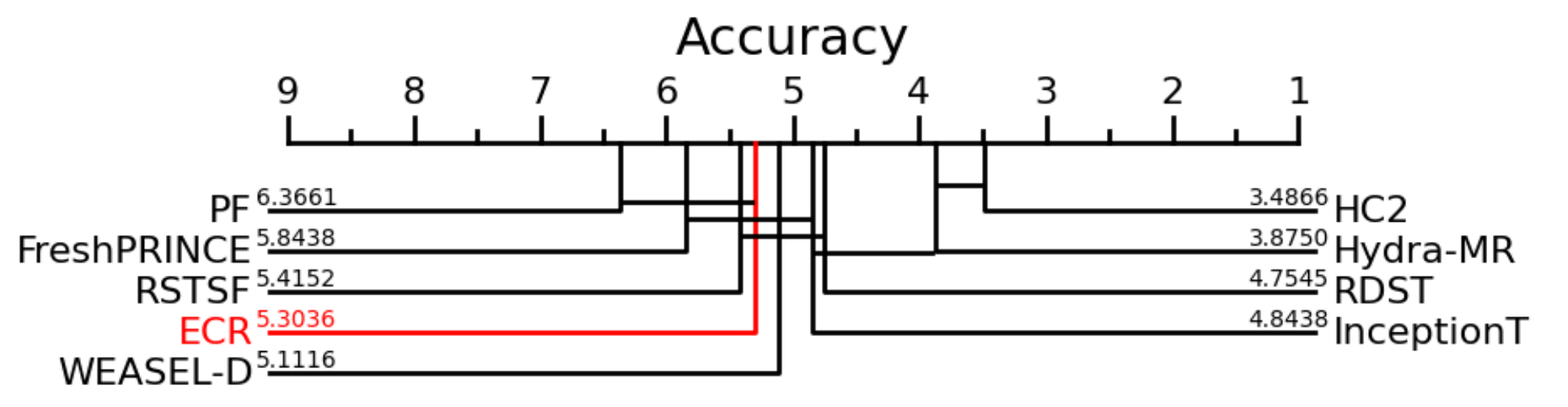}
\caption{Averaged ranked performance statistics for ECR with eight best of category algorithms on 112 UCR UTSC problems.}\label{ECR_vs_sotas}
\end{figure}

\begin{figure*} [htb]
    \centering
    \subfigure[ECRTime versus InceptionT]{
        \label{ECRTime_vs_Inp}
        \includegraphics[width=0.35\linewidth]{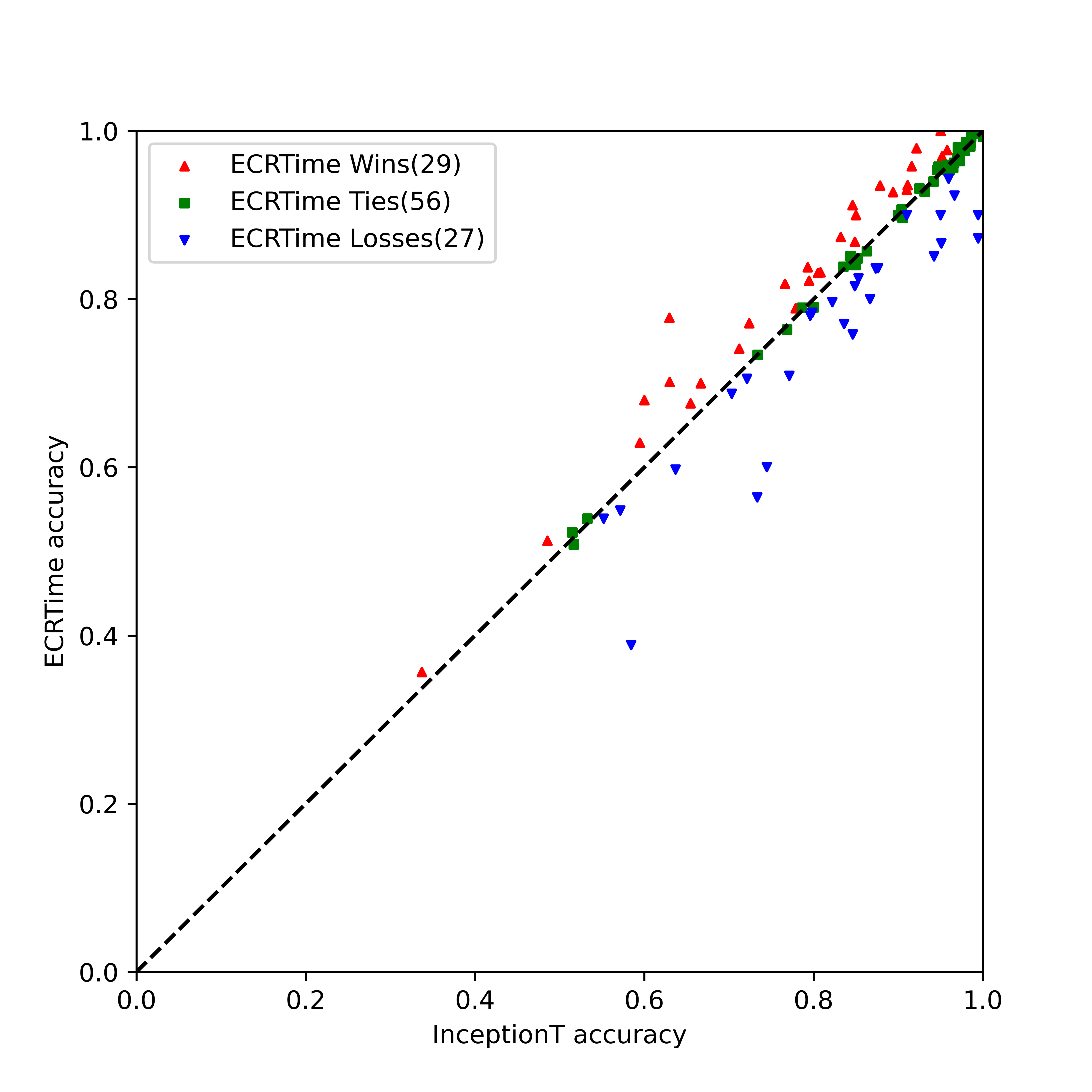}}
    \subfigure[ECRTime versus HIVE-COTE 2.0]{
        \label{ECRTime_vs_HC2}
        \includegraphics[width=0.35\linewidth]{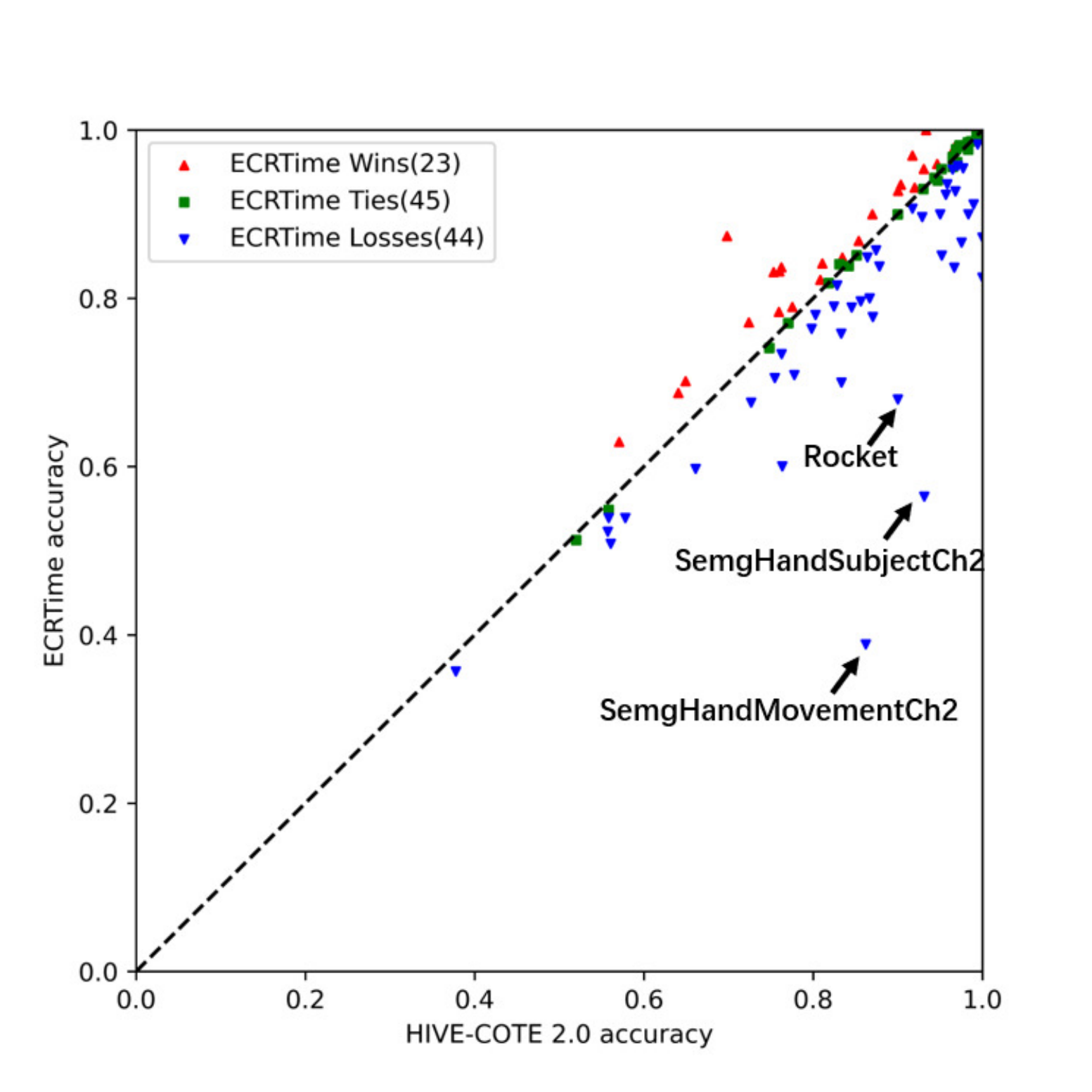}}
    \caption{Scatter charts compare the accuracy of our ECRTime with InceptionT and Hive-COTE 2.0 on 112 UCR datasets.}
    \label{ECRTime_vs_Inp_HC2}
\end{figure*}

In the preceding section, we discussed ECR's significant outperformance of most deep learning classifiers on UCR112. This section broadens the comparative analysis to include all Time Series Classification (TSC) methods. We evaluated ECR against the eight leading state-of-the-art (SOTA) classifiers listed in Backoff-2023\cite{bib27}, namely: (1) HIVE-COTE 2.0\cite{bib11}, (2) Hydra-MR\cite{bib28}, (3) InceptionT\cite{bib13}, (4) RDST\cite{bib29}, (5) WEASEL-D\cite{bib30}, (6) RSTSF\cite{bib31}, (7) FreshPRINCE\cite{bib32}, and (8) PF\cite{bib33}. Of these, HIVE-COTE 2.0 is recognized as the most accurate algorithm for TSC issues, albeit with considerable computational demands. InceptionT, on the other hand, is currently the most accurate deep learning-based TSC classifier. Detailed analysis of these SOTA classifiers can be found on website \footnote{\url{http://timeseriesclassification.com/results.php}}. Adhering to the methodology outlined in Backoff-2023, we conducted training and testing on the original UCR112 train/test set, as depicted in \Cref{ECR_vs_sotas}. The results indicate that ECR outperforms PF (the leading distance-based method), FreshPRINCE (the top feature-based method), and RSTSF (the foremost interval-based method). However, it is surpassed by the other five methodologies.

\begin{figure*}[htbp]
    \centering
    \subfigure[Training time versus length]{
        \label{scalability:a}
        \includegraphics[width=0.4\linewidth]{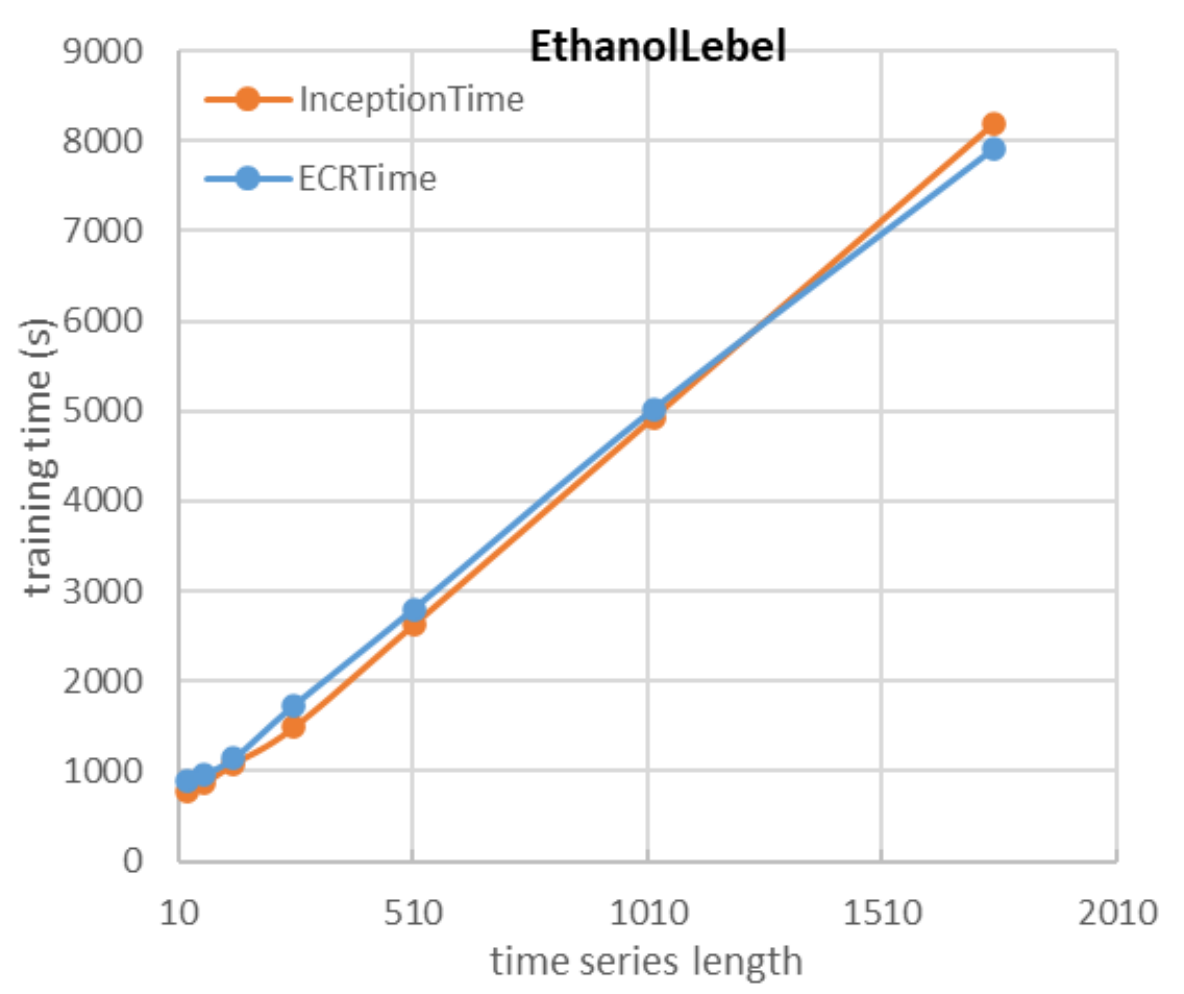}}
    \subfigure[Training time versus size]{
        \label{scalability:b}
        \includegraphics[width=0.45\linewidth]{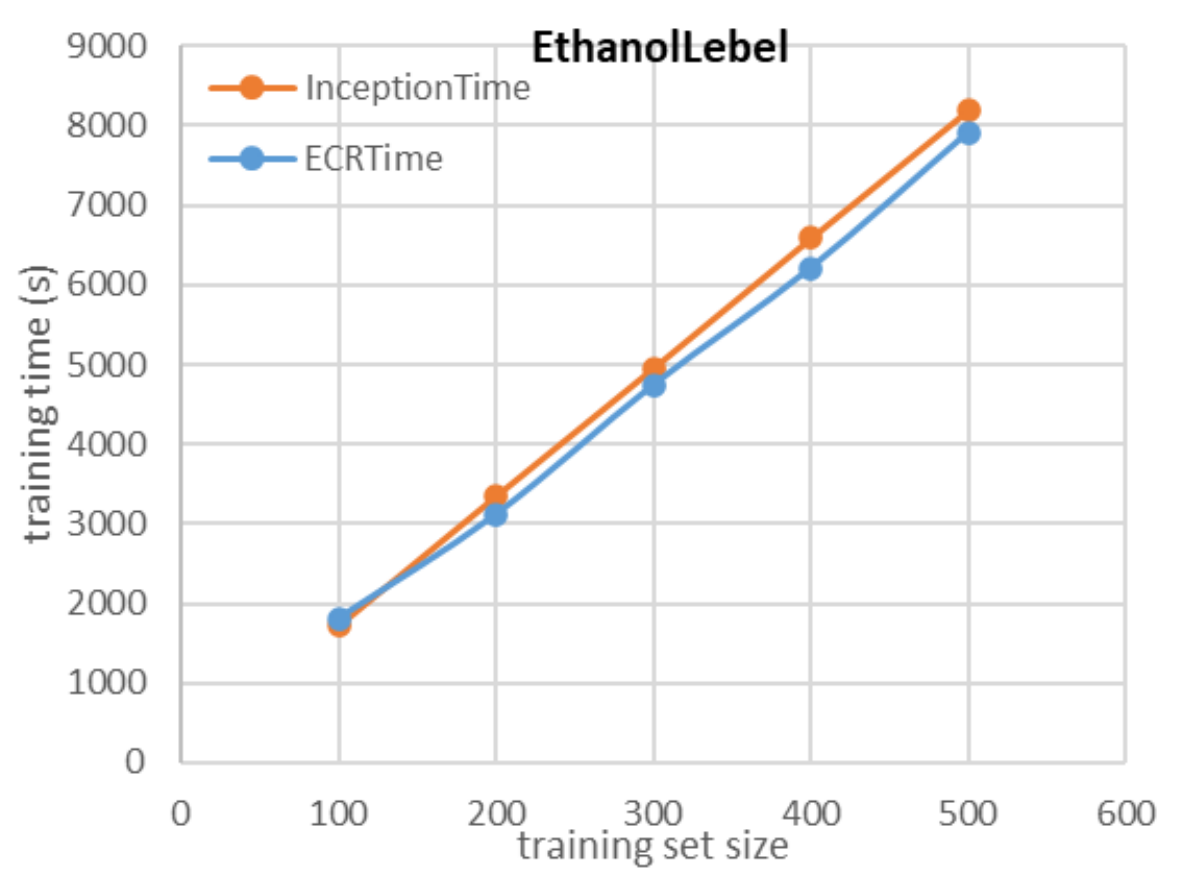}}
    \caption{Training time as a function of the series length and the training set size for the EthanoLebel dataset.}
    \label{scalability}
\end{figure*}

\begin{figure*} [htb]
    \centering
    \subfigure[Distribution of Dataset Types in UCR112 Composition.]{
        \label{Distribution}
        \includegraphics[width=0.45\linewidth]{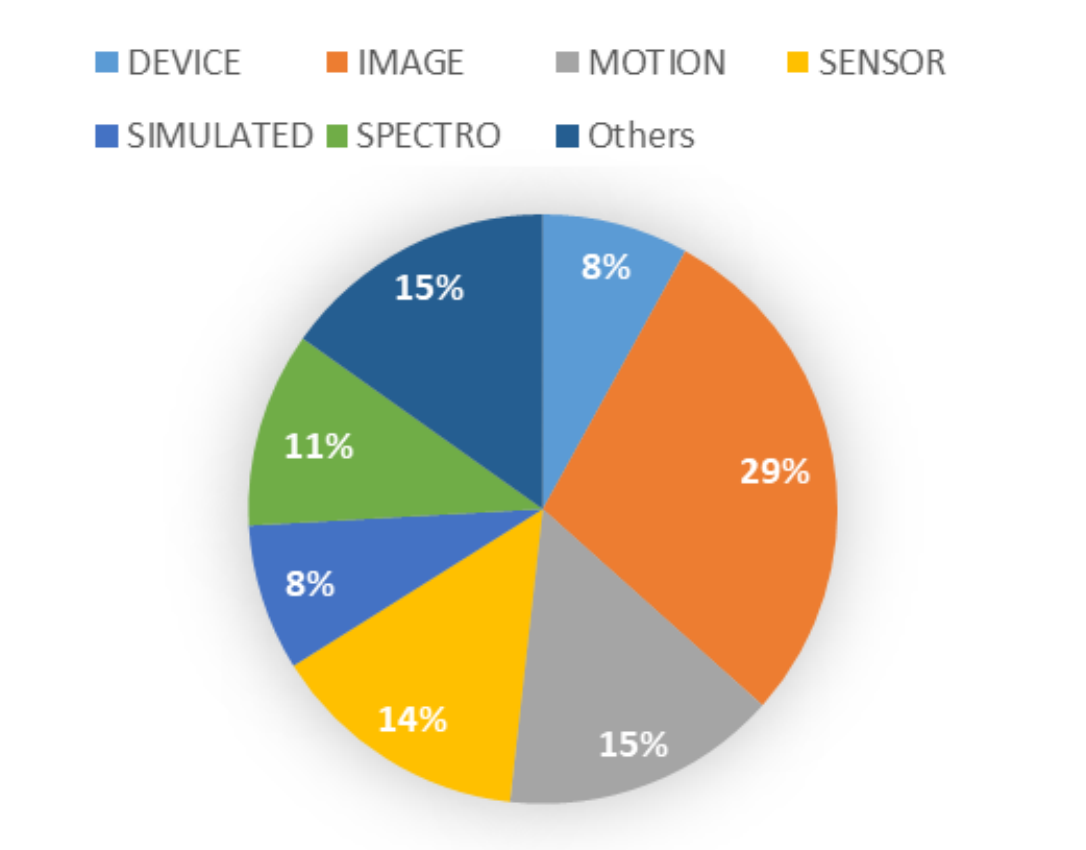}}
    \subfigure[Boxplot Comparison of ECRTime with InceptionT.]{
        \label{Boxplot}
        \includegraphics[width=0.5\linewidth]{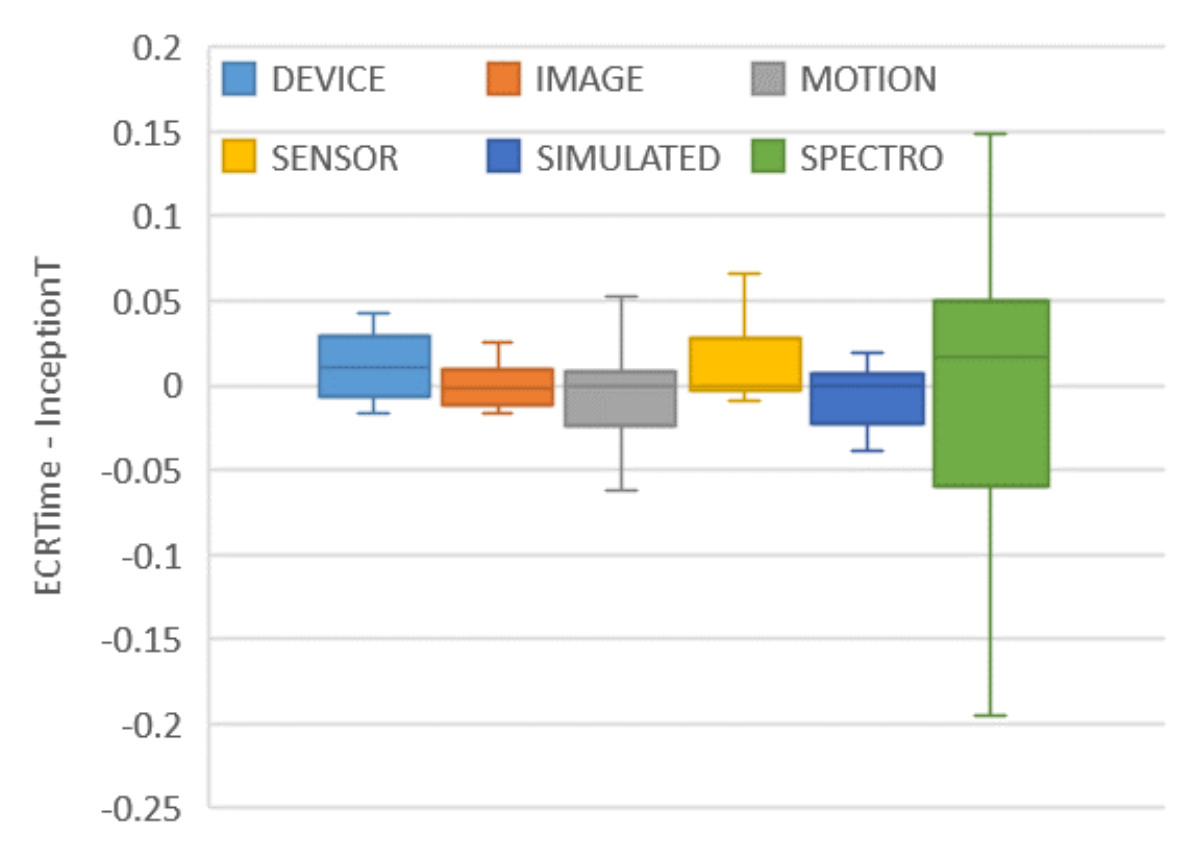}}
    \caption{Comparison of ECRTime and InceptionT based on dataset source types.}
    \label{Distribution_Boxplot}
\end{figure*}

To extend our comparative analysis, we utilized the more efficacious ECRTime, an ensemble of three ECR models, against other SOTA methods. We provided scatter charts for detailed pairwise comparisons: between ECRTime and the most accurate deep learning-based method, InceptionT, as illustrated in \Cref{ECRTime_vs_Inp}, and with the leading algorithm, HIVE COTE 2.0, in \Cref{ECRTime_vs_HC2}. As depicted in \Cref{ECRTime_vs_sotas}, ECRTime marginally outperforms InceptionT, thereby becoming the most precise deep learning classification method currently used in TSC, although its margin over the second-ranked Hydra-MR is not statistically significant (p-value\textgreater 0.05). Nonetheless, it considerably trails behind HIVE-COTE 2.0 (HC2), the most accurate classifier. In \Cref{ECRTime_vs_Inp}, ECRTime and InceptionT demonstrate comparable performance on half of the UCR112, each exhibiting superiority in the remaining datasets, with ECRTime having a narrow advantage (Wins: 29 versus 27). \Cref{ECRTime_vs_HC2} shows a notable gap between ECRTime and HC2 (Wins: 23 versus 44), with ECRTime displaying over a 20\% shortfall in datasets such as Rocket, SemgHandMovementCh2, and SemgHandSubjectCh2. Future efforts will be directed towards enhancing ECRTime's performance in these specific datasets.

Upon synthesizing the comparison results, it becomes clear that the ECRTime model introduced in this study matches the current top-performing deep learning classifier, InceptionT, in terms of overall effectiveness. To conduct a more detailed analysis of their strengths and weaknesses, we compared them based on the dataset types present in UCR112, namely the sequence source types. As depicted in \Cref{Distribution}, UCR112 encompasses 13 dataset types\cite{bib27}. The DEVICE, IMAGE, MOTION, SENSOR, SIMULATED, and SPECTRO categories, which constitute 85\% of the datasets, are the most significant, with DEVICE and SIMULATED being the smallest (9 datasets each) and IMAGE being the largest (32 datasets). The "others" category comprises 7 dataset types, each containing only 1-3 datasets. Due to the dominance of the first six categories in UCR112, we utilized boxplots to illustrate the accuracy variances between ECRTime and InceptionT across these categories. As shown in \Cref{Boxplot}, ECRTime exceeds InceptionT in the DEVICE and SENSOR categories, equals InceptionT in the IMAGE and SPECTRO categories, and is marginally less effective than InceptionT in the MOTION and SIMULATED categories. These insights provide valuable guidance for researchers in choosing the most suitable approaches for diverse practical applications.

\subsection{Runtime analysis}\label{subsection4.4}

\begin{table}[htb]
\caption{Run time to train UCR 112 problems in a single thread. ECRTime was trained on NVIDIA RTX3060(12G) GPU, the other algorithms are reported in Backoff-2023\cite{bib27}.}
    \centering
    \begin{tabular}{ccc}
    \hline
        \textbf{Algorithm} & \textbf{Mean runtime(m)} & \textbf{Platform} \\ \hline
        Hydra-MR & 0.538 & CPU \\ 
        RSTSF & 0.687 & CPU \\ 
        WEASEL-D & 0.818 & CPU \\ 
        RDST & 1.948 & CPU \\ 
        FreshPRINCE & 12.725 & CPU \\ 
        ECRTime & 34.251 & GPU \\ 
        InceptionT & 50.039 & GPU \\ 
        HIVE-COTE 2.0 & 271.168 & CPU \\ 
        PF & 381.897 & CPU \\ \hline
    \end{tabular}
    \label{Runtime}
\end{table}

To enable a comparison of time efficiency with ECRTime, we extracted the average training time of various state-of-the-art methods on UCR142 from Backoff-2023 as an approximate for the time spent on UCR112. These comparative findings are presented in \Cref{Runtime}. ECRTime exhibits a reduced training duration compared to InceptionT. While it is more time-intensive than other CPU-based methods like RDST and Hydra-MR, leveraging parallel training on multiple GPUs can decrease its training time, as ECRTime is GPU-based. It is important to note that for this study, ECRTime was trained using an RTX3060 GPU, a less powerful consumer-grade graphics card. Utilizing more advanced graphics cards could lead to a further decrease in training duration.

\begin{figure}[htp]%
\centering
\includegraphics[width=0.48\textwidth]{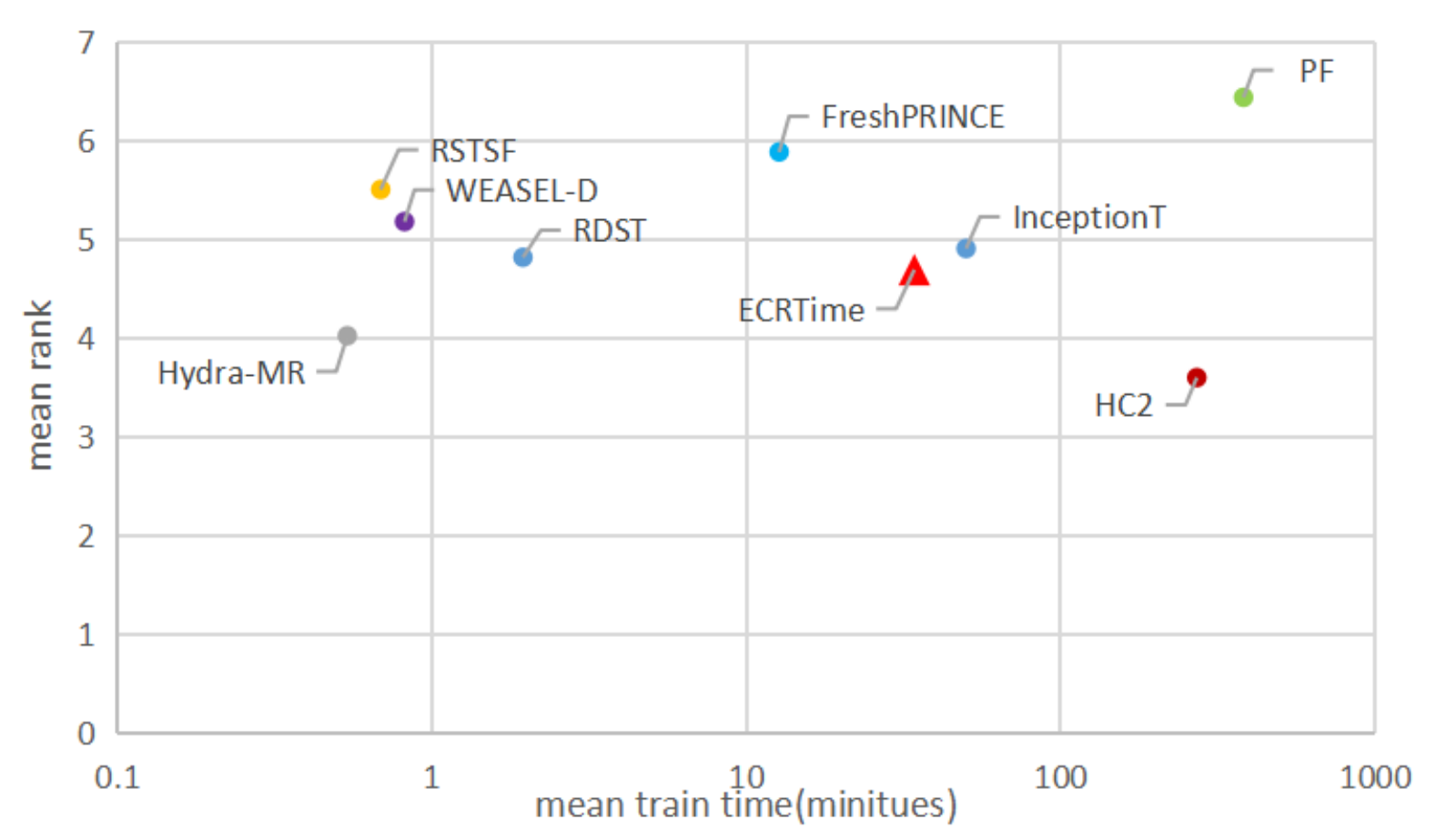}
\caption{Mean rank versus mean training time for each state-of-the-art method. The training time is represented on a log scale.}\label{time_vs_rank}
\end{figure}

For a comprehensive comparison of accuracy and training duration across various methods, we utilized the average training time from \Cref{Runtime} as the X-axis and the mean rank on the UCR112 dataset as the Y-axis to construct \Cref{time_vs_rank}. This figure indicates that Hydra-MR, a member of the ROCKET family, presents the most favorable balance between accuracy and training time. HC2 achieves the highest accuracy but requires the longest training period, whereas ECRTime is ranked third in accuracy but benefits from a relatively short training duration. To examine the scalability of ECRTime and InceptionT in depth, we assessed the relationship between training time and sequence length, as well as training time and the number of samples in the training set, using the EthanolLevel dataset from UCR112. As illustrated in \Cref{scalability:a} and \Cref{scalability:b}, ECRTime demonstrates a slightly more gradual increase in training duration compared to InceptionT, in response to longer sequence lengths or larger training sample sizes. Consequently, when considering factors such as training time, accuracy, and scalability, ECRTime emerges as a more advantageous option compared to InceptionT.

\subsection{Sensitivity study}\label{subsec4.5}

We explore the effect of key parameter choices on accuracy over UCR112 for ECR and ECRTime:

\begin{itemize}
\item[--] 1-NN classifier versus SoftMax classifier.

\item[--] hard triplet loss versus triplet loss.

\item[--] ensemble in ECR.

\item[--] ensemble in ECRTime.

\item[--] batch size.
\end{itemize}


\begin{figure*} [ht]
    \centering
    \subfigure[1-NN versus SoftMax]{
        \label{classifier}
        \includegraphics[width=0.35\linewidth]{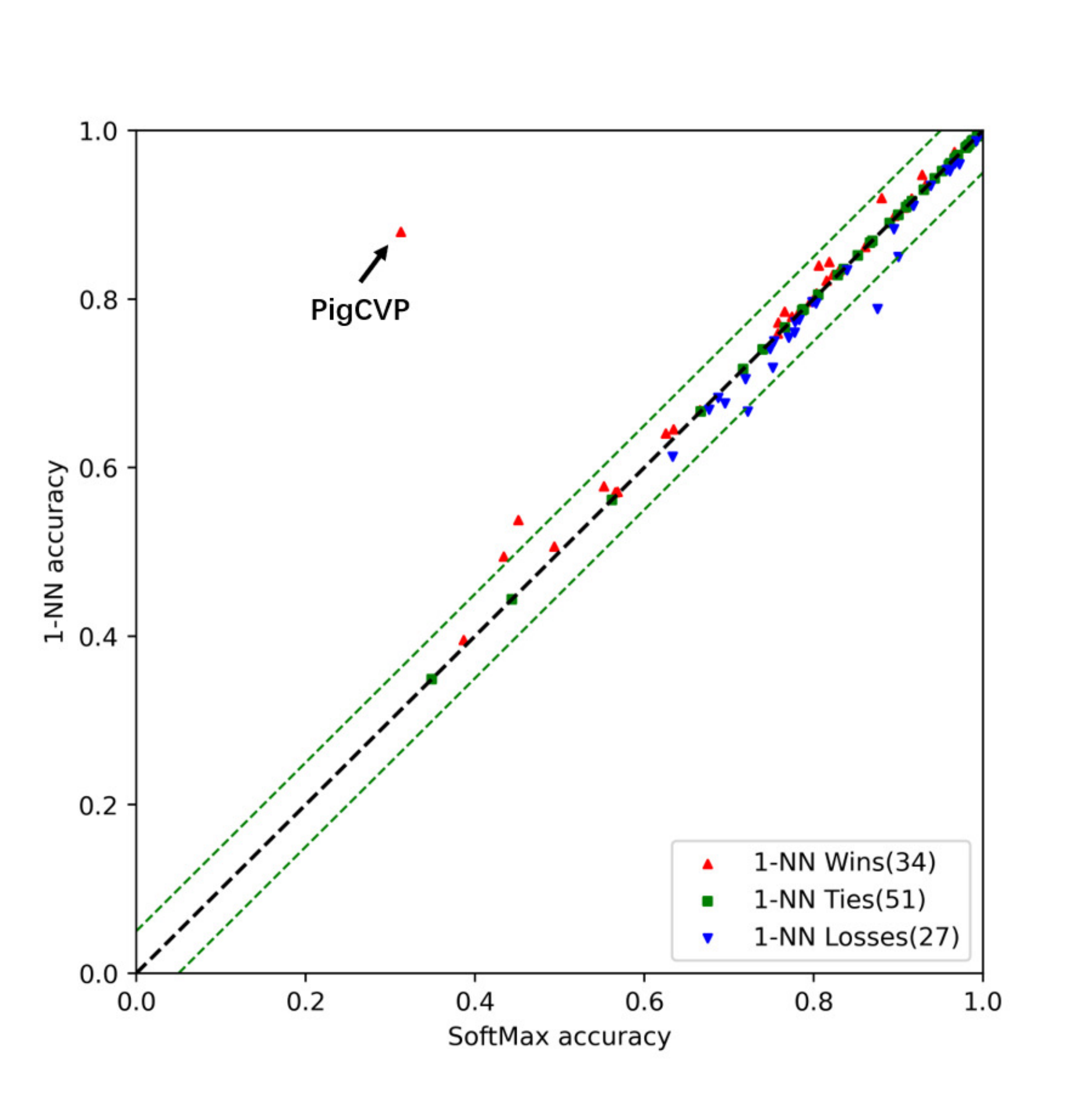}}
    \subfigure[Hardtriplet versus triplet]{
        \label{loss}
        \includegraphics[width=0.35\linewidth]{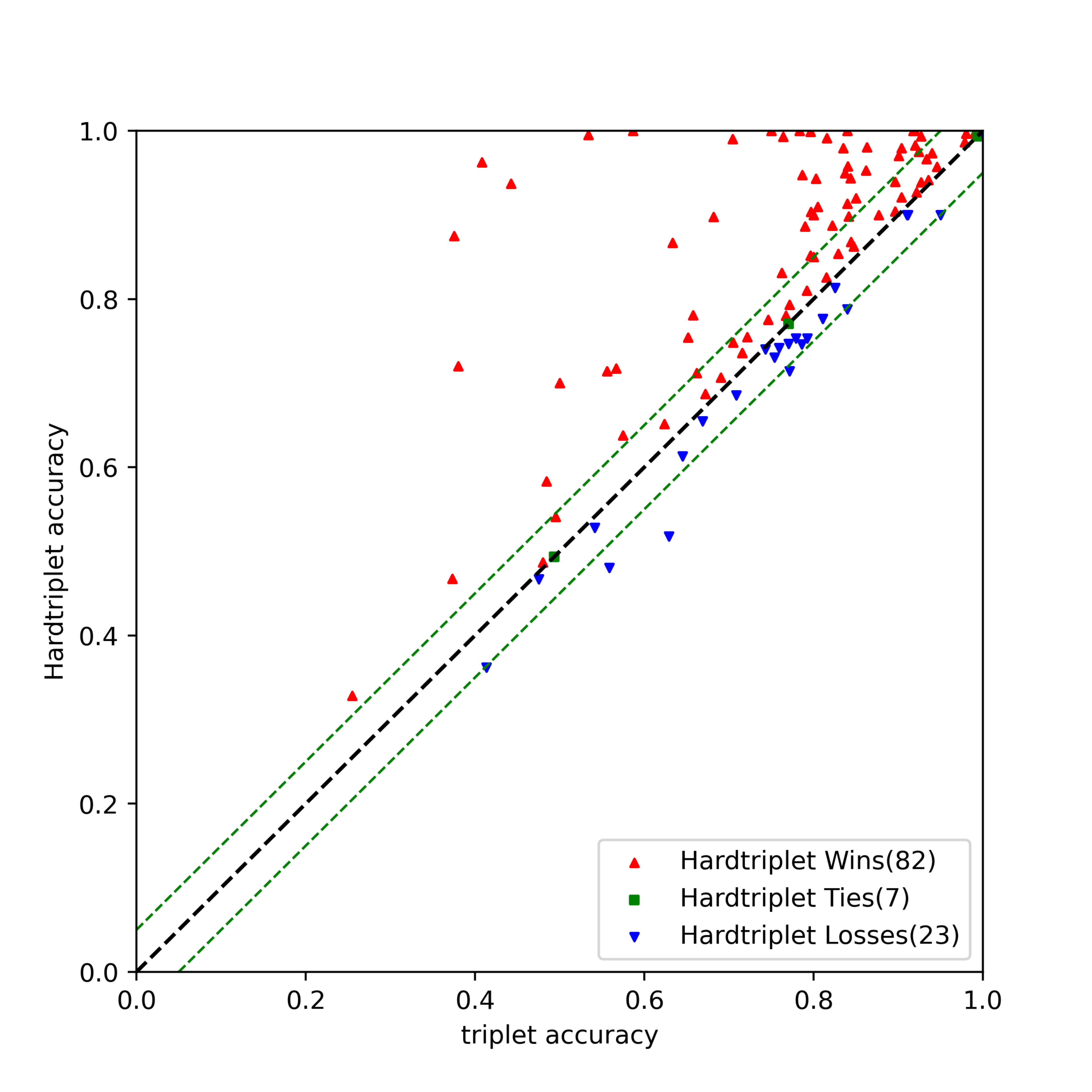}}
    \caption{Pair scatter plots for classifiers and losses. Each point represents the accuracy of each dataset. The dotted lines indicate ±5\% interval on the classification accuracy.}
    \label{classifier_loss}
\end{figure*}

\subsubsection{1-NN classifier versus SoftMax classifier}\label{subsubsec4.5.1}

\begin{figure*} [ht]
    \centering
    \subfigure[retrieval versus classification]{
        \label{ECR_ensemble_1}
        \includegraphics[width=0.3\linewidth]{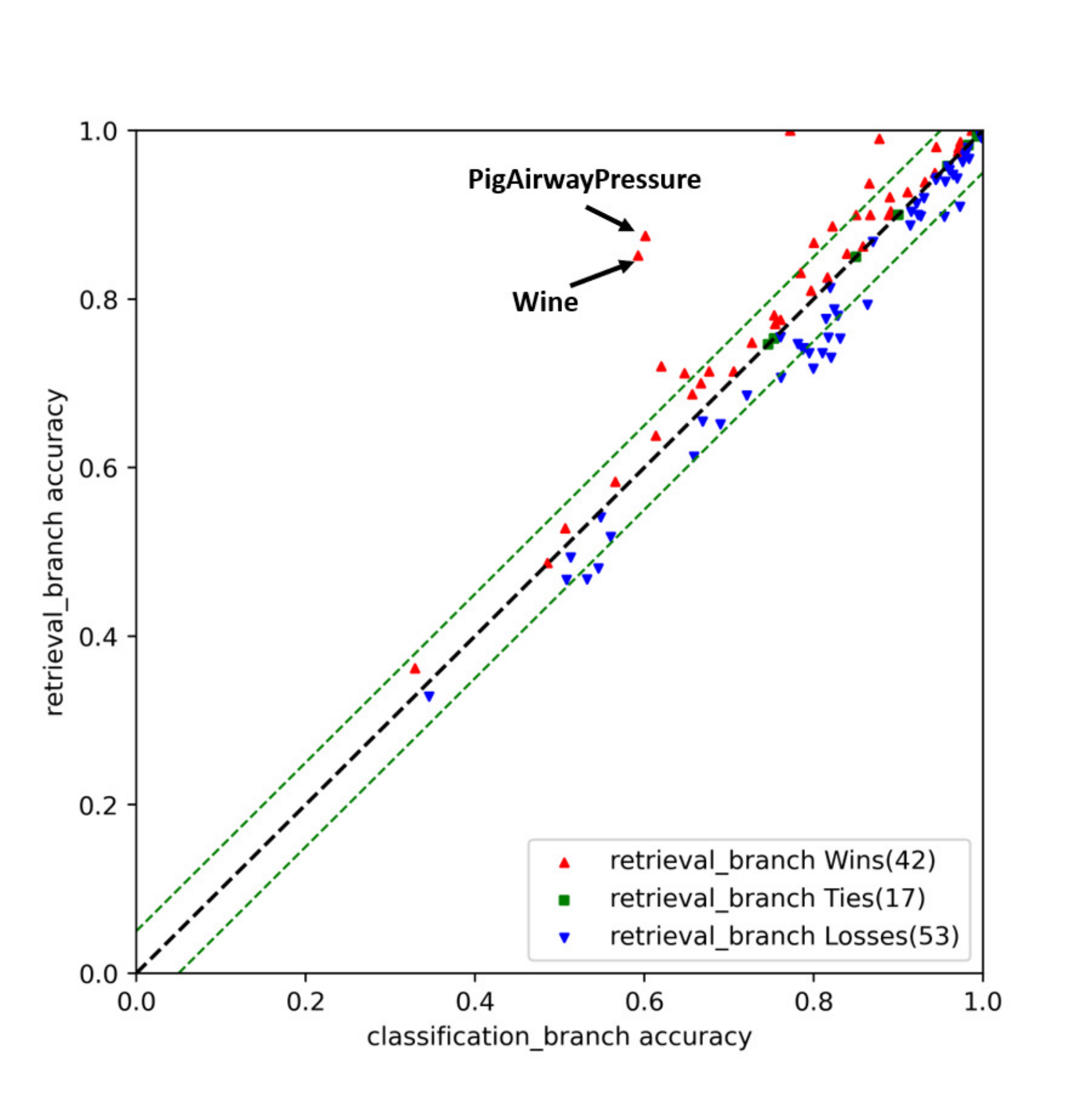}}
    \subfigure[ECR versus classification]{
        \label{ECR_ensemble_2}
        \includegraphics[width=0.3\linewidth]{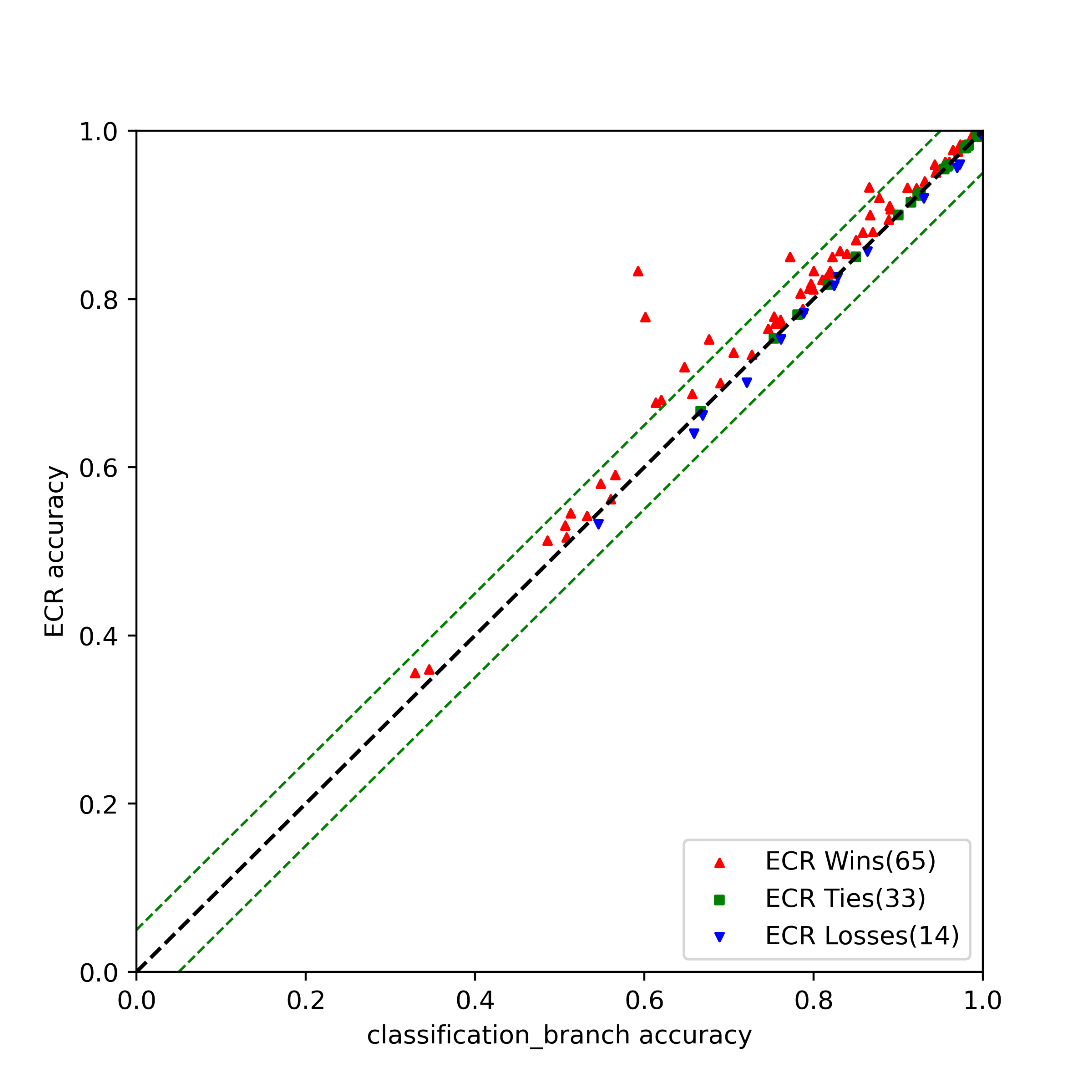}}
    \subfigure[ECR versus retrieval]{
        \label{ECR_ensemble_3}
        \includegraphics[width=0.3\linewidth]{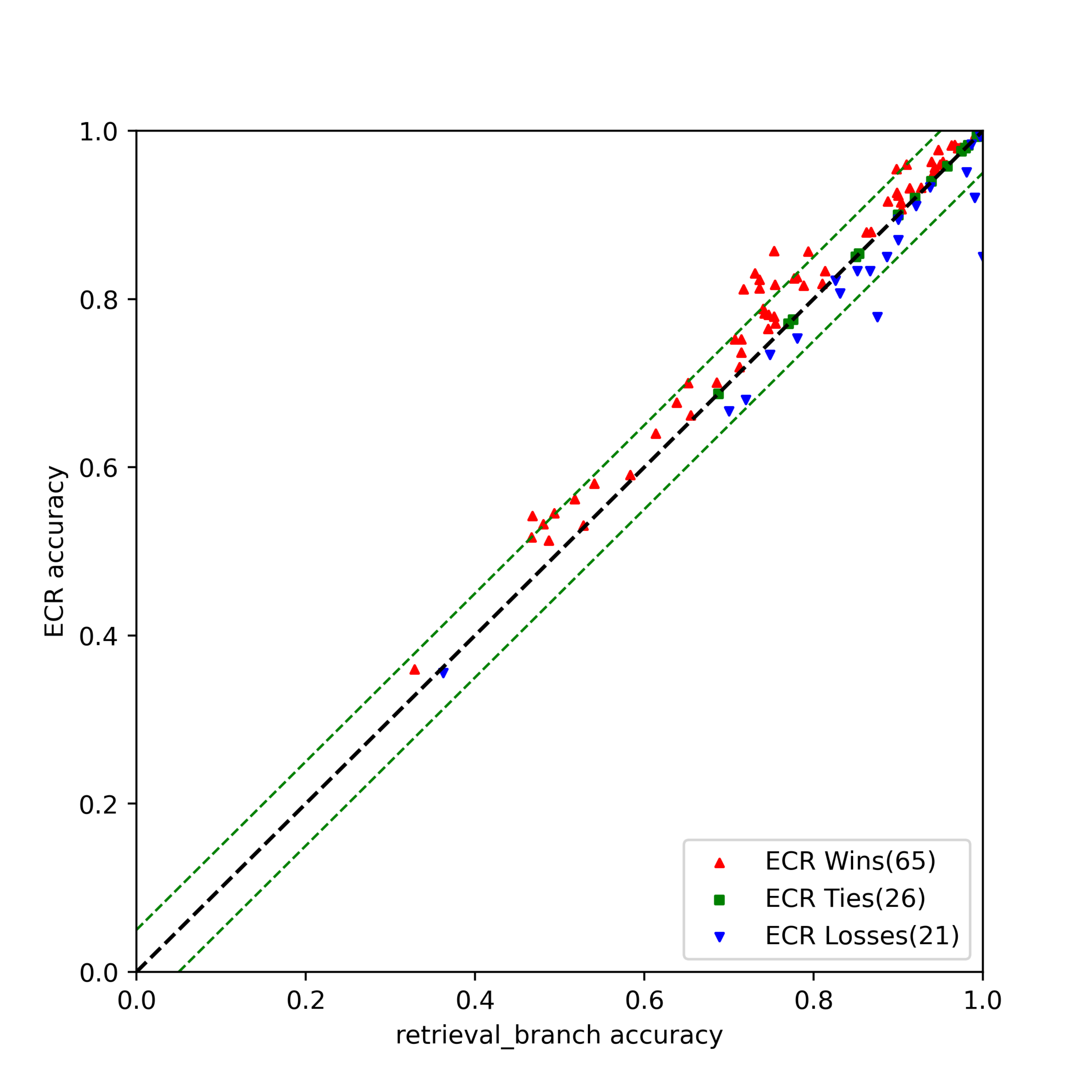}}
    \caption{Pair scatter plots for ensemble in ECR. Each point represents the average accuracy value over 112 datasets. The dotted lines indicate ±5\% interval on the accuracy.}
    \label{ECR_ensemble}
\end{figure*}

\begin{figure*} [htp]
    \centering
    \subfigure[ECR versus classification(2)]{
        \label{ECR_classification2}
        \includegraphics[width=0.3\linewidth]{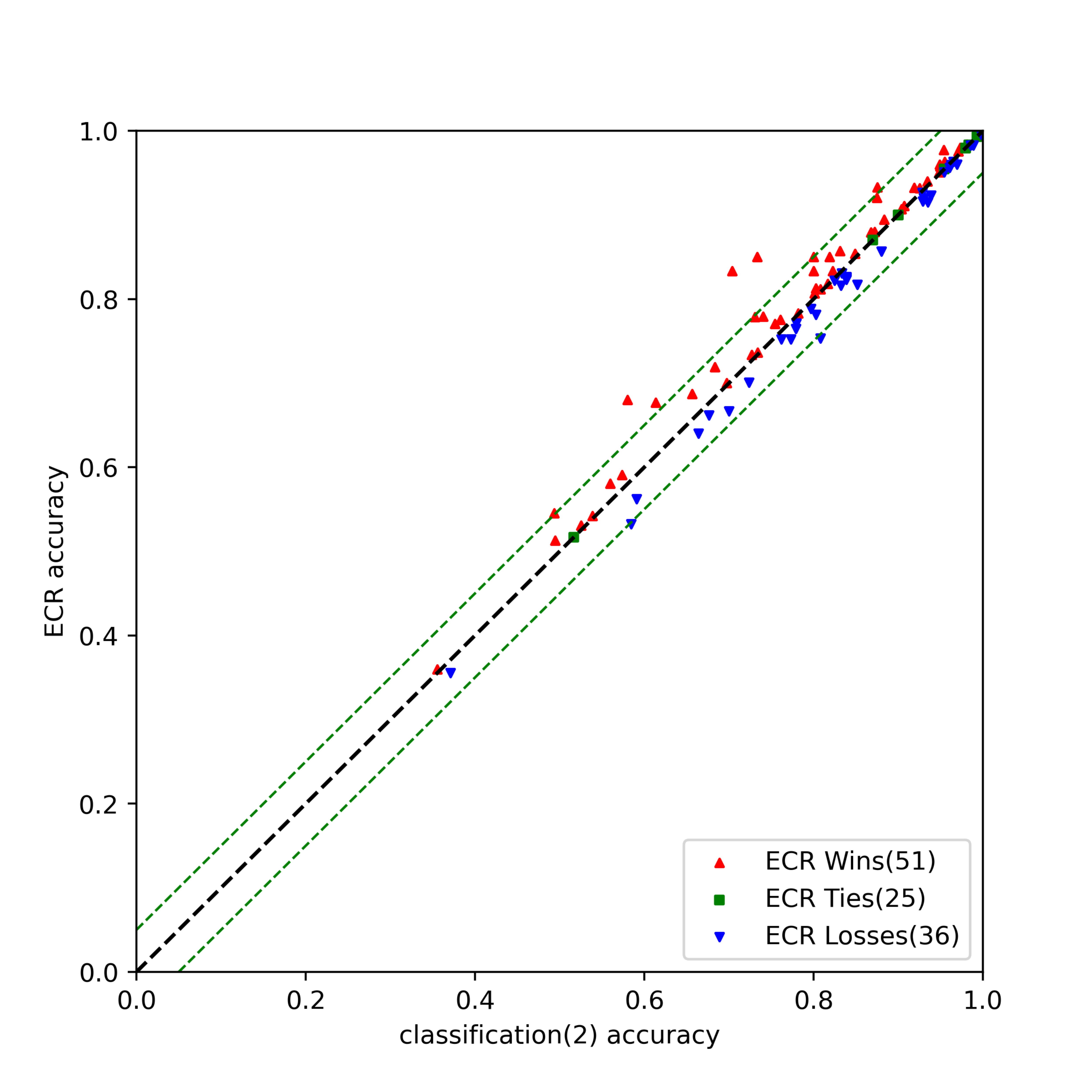}}
    \subfigure[ECR versus retrieval(2)]{
        \label{ECR_retrieval2}
        \includegraphics[width=0.3\linewidth]{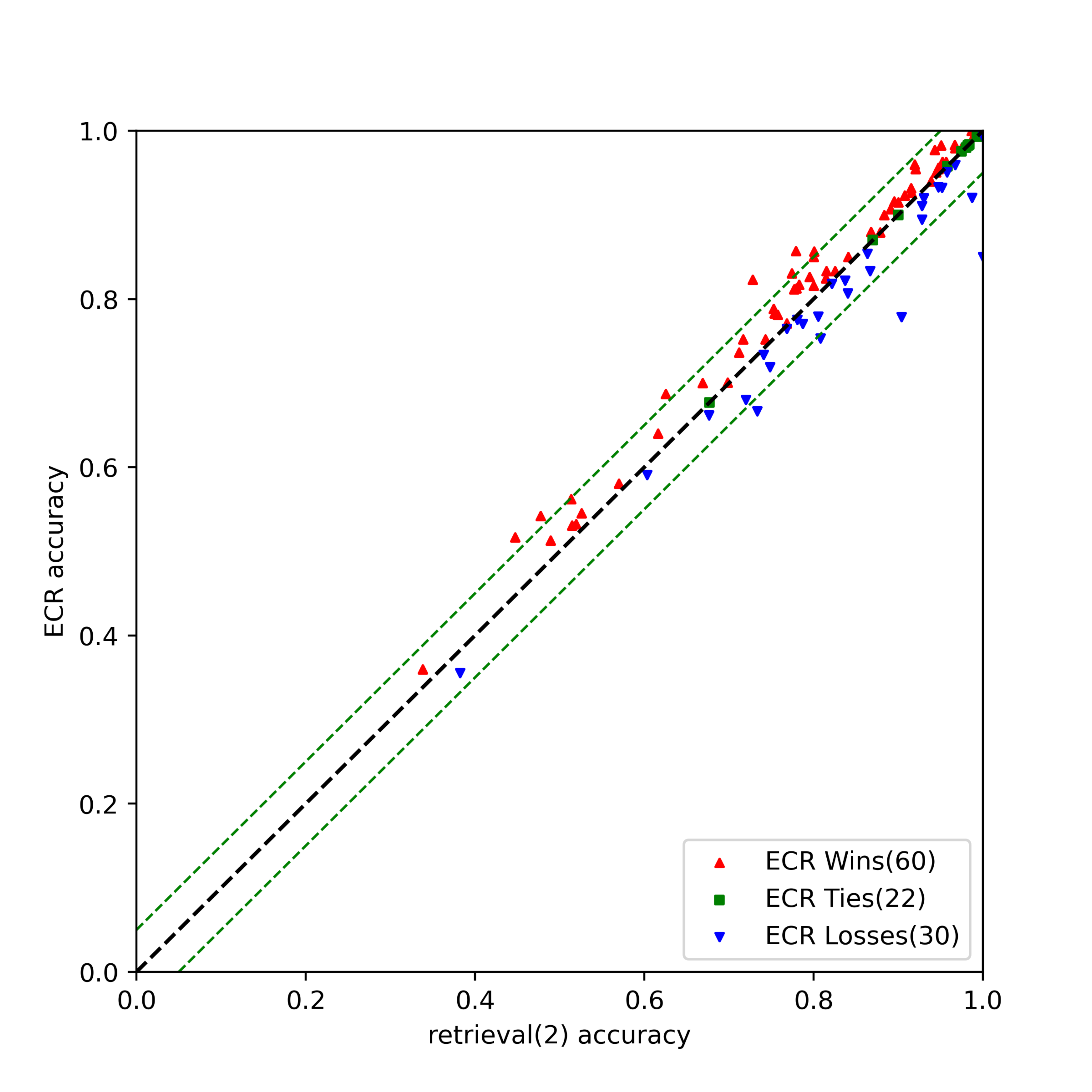}}
    \subfigure[ECRTime versus ECR]{
        \label{ECRTime_vs_ECR}
        \includegraphics[width=0.3\linewidth]{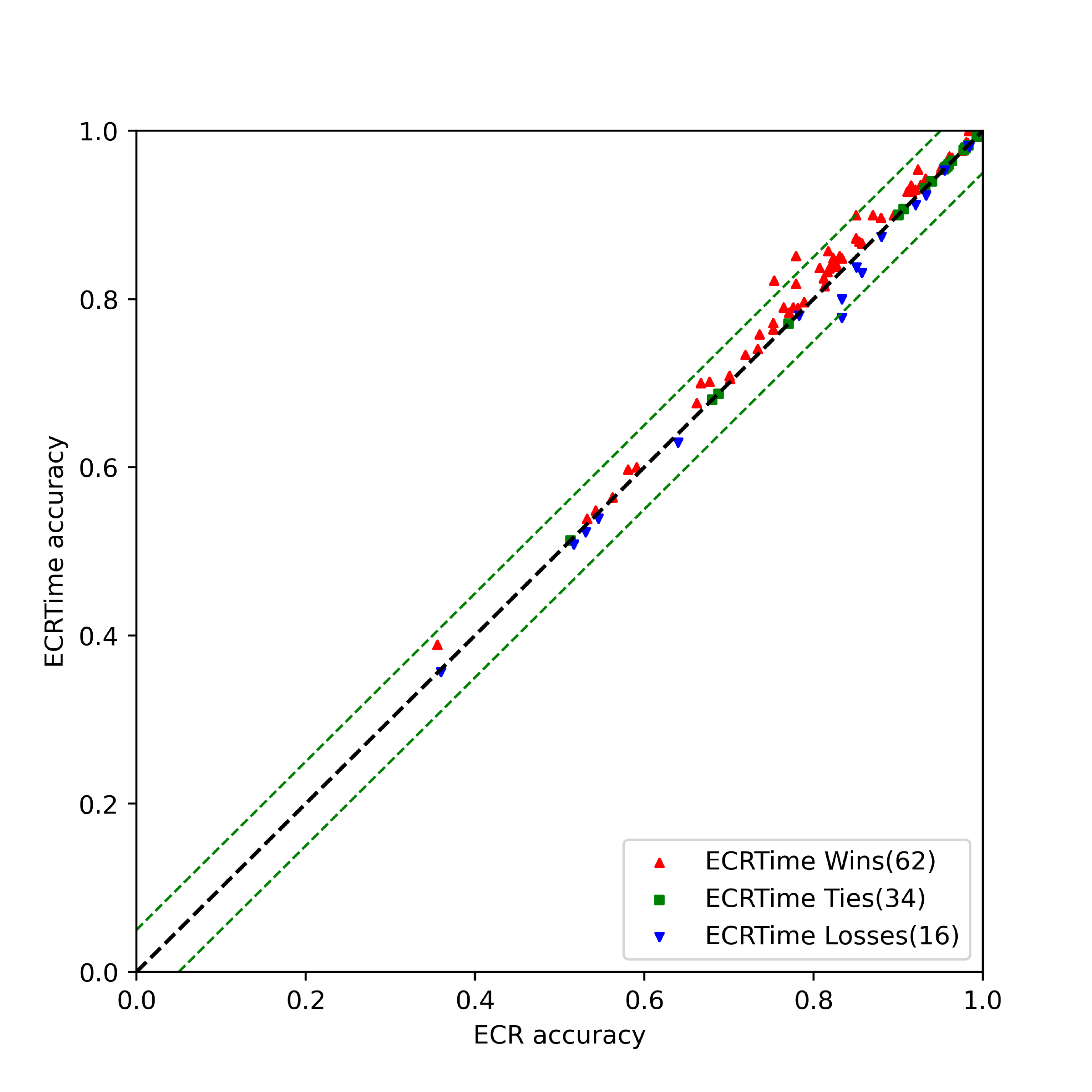}}
    \caption{Pair scatter plots of ECR with classification(2), retrieval(2) and ECRTime. classification(2) means an ensemble of two classification modules. The dotted lines indicate ±5\% interval on the classification accuracy.}
    \label{ECR_ECRTime_ensemble}
\end{figure*}

In Section~\ref{sec1}, the phenomenon of ``inter-class similarity and intra-class inconsistency'' within UCR datasets was explored, and its adverse effect on the SoftMax classifier was analyzed. Consequently, the implementation of a 1-NN classifier was suggested as a potential mitigation strategy. This subsection details comparative experiments conducted on UCR112, contrasting the 1-NN classifier with the SoftMax classifier, based on ECR's classification sub-model. As depicted in \Cref{classifier}, the integration of a 1-NN classifier with the classification network backbone shows a marginally better performance than the ``FC+SoftMax'' approach, though the difference is not statistically significant (p-value\textgreater 0.05). Specifically, the average accuracies of these two methods on UCR112 were calculated to be 84.04\% and 83.52\%, respectively, signifying a modest improvement of 0.5\% with the 1-NN classifier. Remarkably, for the HEMODYNAMICS-type PigCVP dataset, accuracy using the SoftMax classifier was a mere 31.25\%, which notably increased to 87.98\% when employing the 1-NN classifier, offering insightful implications for practical applications.

\subsubsection{hard triplet loss versus triplet loss}\label{subsubsec4.5.2}

To assess the impact of hard triplet loss versus triplet loss on the model, we separately trained the retrieval sub-model of ECR on UCR112 using each loss type. The findings, as illustrated in \Cref{loss}, indicate that hard triplet loss significantly enhances performance compared to standard triplet loss. Notably, the model employing hard triplet loss surpasses the latter in 82 of the 112 datasets, often by margins exceeding 5\%, and with the greatest improvement approaching 50\%. Furthermore, while the model underperforms relative to the triplet loss in 23 datasets, the performance decrease generally remains below 5\%. These outcomes establish hard triplet loss as a considerably more effective option than traditional triplet loss.

\subsubsection{batch size}\label{subsubsec4.5.3}

\begin{figure}[htp]%
\centering
\includegraphics[width=0.48\textwidth]{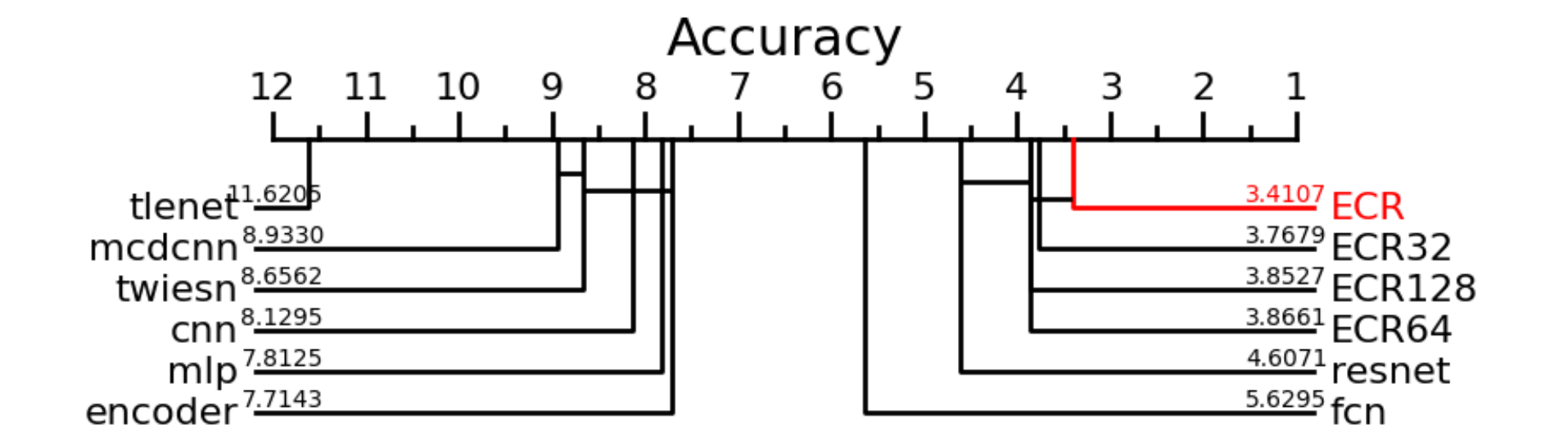}
\caption{Critical difference diagram showing the effect of the batch size hyperparameter value over the average rank of ECR. ECRn denotes ECR with a batch size set to n, with the default being 16.}\label{batchsize}
\end{figure}

The critical difference diagram presented in \Cref{batchsize} elucidates the effect of batch size on ECR's performance. A horizontal line across different models in the diagram suggests no substantial difference in their performance across the 112 datasets, with a slight advantage for ECR(batch size equal to 16). Additionally, the diagram indicates that at batch sizes of 64 and 128, ECR does not demonstrate a significant benefit over ResNet (p-value\textgreater 0.05). ECR's performance markedly surpasses that of other deep learning methods only at batch sizes of 32 and 16 (p-value\textless 0.05). In this study, the chosen default batch size for ECR is 16.

\subsubsection{ensemble in ECR}\label{subsubsec4.5.4}

\begin{figure}[htp]%
\centering
\includegraphics[width=0.48\textwidth]{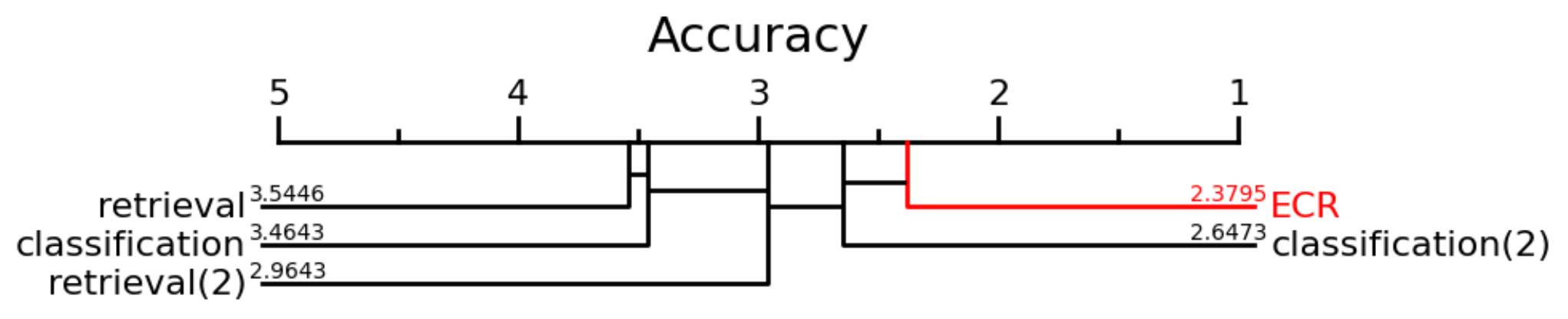}
\caption{Mean rank of classification module, retrieval module, classification(2) and retrieval(2) in terms of accuracy versus ECR over 112 datasets from the UCR archive. classification(2) means an ensemble of two classification modules.}\label{ECR_ensemble_rank}
\end{figure}

\Cref{ECR_ensemble} shows the pairwise comparison results of the retrieval module vs. classification module, ECR vs. classification module, and ECR vs. retrieval module. As depicted in \Cref{ECR_ensemble_1}, a parity in performance is observed in only 17 of the 112 datasets, while each module exhibits strengths in the remaining 95 datasets. Remarkably, the retrieval module secures a performance edge exceeding 20\% in datasets such as PigAirwayPressure and Wine. To harness the benefits of both modules, we integrated the classification and retrieval submodules to form the ECR model. This integration's efficacy, as demonstrated in \Cref{ECR_ensemble_2}, \Cref{ECR_ensemble_3}, and \Cref{ECR_ensemble_rank}, reveals that the composite ECR model surpasses the performance of each individual submodule, thereby confirming the ensemble strategy's effectiveness. Additionally, an evaluation of ECR in contrast to classification(2) and retrieval(2), detailed in \Cref{ECR_ECRTime_ensemble}, establishes that ensembling two classification submodules or two retrieval submodules is less effective compared to an ensemble comprising one of each. Classification(2) refers to the approach of ensembling two classification models, following the method outlined in Eq.~\eqref{eq6}. Similarly, retrieval(2) follows a comparable approach. This finding underscores the complementary nature of the submodules within ECR, effectively balancing their respective strengths and limitations.

\subsubsection{ensemble in ECRTime}\label{subsubsec4.5.5}

\begin{figure}[htp]%
\centering
\includegraphics[width=0.48\textwidth]{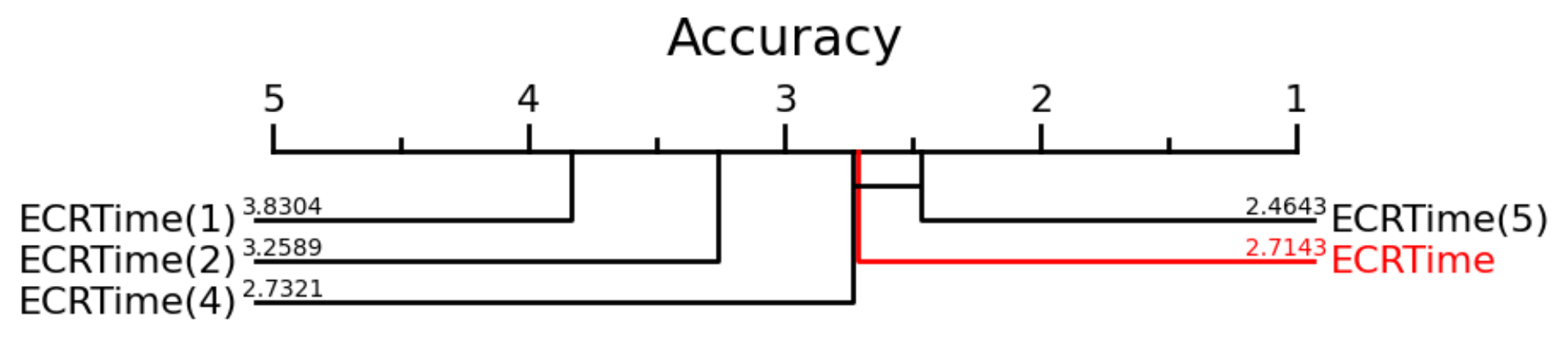}
\caption{Critical difference diagram showing the effect of the ensemble num hyperparameter value over the average rank of ECRTime. ``ECRTime(n)'' notation is used to denote an ensemble of n ECR modules, with the default being 3.}\label{ensemble_num}
\end{figure}

The final ECRTime model presented in this study, which attains enhanced performance through the integration of multiple ECRs, was subjected to a comparative analysis focusing on the number of modules in the ensemble as a key hyperparameter. \Cref{ensemble_num} illustrates that the performance of ECRTime significantly increases when the number of ECRs increases from 1 to 3. However, further expansion to 4 and 5 does not yield a notable improvement in performance, while concurrently increasing training duration. To strike an optimal balance between accuracy and computational efficiency, thereby boosting practical usability, this study finalizes the ensemble at three modules.

A detailed examination of the enhancements achieved by ensembling three ECR models is conducted through a pairwise comparison between ECRTime and ECR on UCR112, employing scatter charts as presented in \Cref{ECRTime_vs_ECR}. This figure reveals that the ensembled ECRTime model exhibits improvements in 62 of the 112 datasets, though these improvements predominantly fall within a 5\% range. A marginal decrease in performance is observed in 16 datasets, while the remaining 34 datasets exhibit no variation. Furthermore, a post-hoc statistical analysis confirms a significant distinction between ECRTime and ECR (p-value\textless 0.05). In conclusion, ECRTime demonstrates better performance than ECR in time series classification tasks.


\section{Conclusion}\label{sec5}

In the domain of deep learning-based time series classification employing the ``FC+SoftMax'' paradigm, replacing the SoftMax classifier with a 1-NN classifier has resulted in enhanced performance. Furthermore, to explicitly adapt to the classification objectives of the 1-NN classifier, we innovatively introduce a deep learning-based retrieval method for TSC issues. By combining this with the classification model in an ensemble, we present the ECRTime framework in this paper.

ECRTime exhibits a highly competitive and advanced standard in terms of accuracy and time complexity, matching or surpassing current state-of-the-art (SOTA) methods in Time Series Classification (TSC) tasks. In future research, we aim to delve deeper into the potential applications of retrieval methods within time series classification and to expand our exploration into the realm of multi-dimensional time series classification.



\bibliographystyle{elsarticle-num} 
\bibliography{elsarticle-bibliography}





\end{document}